\documentclass[runningheads]{llncs}
\usepackage{multirow}
\usepackage{pifont} 
\newcommand{\cmark}{\ding{51}} 
\newcommand{\xmark}{\ding{55}} 
\usepackage{wrapfig}
 
\usepackage{eccv}

\usepackage{eccvabbrv}

\usepackage{graphicx}
\usepackage{booktabs}

\usepackage[accsupp]{axessibility}  
\usepackage{xcolor}
\newcommand{\best}[1]{\textbf{\textcolor{red}{#1}}}
\newcommand{\secondbest}[1]{\underline{\textcolor{blue}{#1}}}
\newcommand{\third}[1]{\textcolor{ForestGreen}{\textit{#1}}}
\usepackage{multirow}
\usepackage{nicematrix}
\usepackage{bbm}
\usepackage{makecell}
\newcommand\shline{\specialrule{1.0pt}{0pt}{0pt}}

\usepackage{hyperref}

\usepackage{orcidlink}

\begin{document}

\title{OARS: Process-Aware Online Alignment for Generative Real-World Image Super-Resolution} 

\titlerunning{OARS}

\author{
\parbox[t]{0.95\textwidth}{\centering
Shijie Zhao\textsuperscript{*}\inst{1}, Xuanyu Zhang\textsuperscript{*}\inst{1,2}, Bin Chen\inst{1,2},\\
Weiqi Li\inst{1,2}, Qunliang Xing\inst{1}, Kexin Zhang\inst{1}, Yan Wang\inst{1},\\
Junlin Li\inst{1}, Li Zhang\inst{1}, Jian Zhang\inst{2}, and Tianfan Xue\inst{3}
}
}

\authorrunning{S. Zhao, X. Zhang et al.}

\institute{
ByteDance Inc. \and
Peking University \and
The Chinese University of Hong Kong
}

\maketitle



\begingroup
\renewcommand\thefootnote{*}
\footnotetext{Equal contribution.}
\endgroup

\begin{abstract}
Aligning generative real-world image super-resolution models with human visual preference is challenging due to the perception--fidelity trade-off and diverse, unknown degradations. Prior approaches rely on offline preference optimization and static metric aggregation, which are often non-interpretable and prone to pseudo-diversity under strong conditioning. We propose \textbf{OARS}, a process-aware online alignment framework built on \textbf{COMPASS}, a MLLM-based reward that evaluates the \emph{LR$\rightarrow$SR} transition by jointly modeling fidelity preservation and perceptual gain with an input-quality-adaptive trade-off. To train COMPASS, we curate COMPASS-20K spanning synthetic and real degradations, and introduce a three-stage perceptual annotation pipeline that yields calibrated, fine-grained training labels. Guided by COMPASS, OARS performs progressive online alignment from cold-start flow matching to full-reference and finally reference-free RL via shallow LoRA optimization for on-policy exploration. Extensive experiments and user studies demonstrate consistent perceptual improvements while maintaining fidelity, achieving state-of-the-art performance on Real-ISR benchmarks.
\keywords{Super-Resolution \and Reinforcement Learning}
\end{abstract}

\section{Introduction}
\label{sec:intro}

Real-World Image Super-Resolution (Real-ISR) \cite{cai2019toward,wang2021real} is a fundamental problem in low-level vision, aiming to restore high-fidelity and perceptually pleasing high-resolution (HR) images from low-resolution (LR) counterparts suffering from complex, unknown degradations. While early CNN-based methods \cite{dong2014learning,lim2017enhanced} focused on optimizing pixel-wise reconstruction errors, they often produce overly smooth textures that correlate poorly with human visual perception. The advent of Diffusion Models \cite{wang2024exploiting, yue2023resshift} has shifted the paradigm from regression to generative distribution matching, significantly improving perceptual quality. However, standard Supervised Fine-Tuning (SFT) on fixed LR-HR pairs faces inherent limitations: it struggles to generalize to unseen real-world degradations and lacks a direct optimization mechanism to align generated content with human aesthetic preferences, often leading to either hallucinations or over-smoothing artifacts.

Recently, Reinforcement Learning from Human Feedback (RLHF) \cite{zheng2025diffusionnft,xu2023imagereward,wu2025editreward, liu2025flow} has emerged as a powerful technique to align generative models in text and image tasks~\cite{guo2025deepseek,liu2025visual,li2025qinsight}. Despite its potential, applying RL to Real-ISR presents unique challenges, bottlenecked by the reward designs and the optimization mechanisms.

First, while leveraging Image Quality Assessment (IQA) models as RL reward signals is intuitive, directly applying existing metrics presents some hurdles. Full-Reference (FR) \cite{ding2020image,wang2004image} metrics are restricted by the absence of ground-truth (GT) in real-world scenarios. Meanwhile, previous No-Reference (NR) metrics\cite{wang2023exploring,li2025qinsight,zhao2025reasoning} are primarily trained on natural images or generic synthetic degradations, severely lacking the fine-grained sensitivity required to distinguish subtly different generative SR outputs. Furthermore, how to effectively integrate FR and NR metrics into a unified reward remains under-explored. Simply combining these conflicting metrics via static linear summation imposes a rigid optimization objective that ignores the initial degradation severity. Without a dynamic scheme, such static aggregation can potentially lead to either insufficient enhancement or unnatural local over-sharpening.

Second, current offline RL frameworks suffer from pseudo-diversity and exploration collapse. Methods like DP$^2$O-SR \cite{wu2025dp2osr} construct preference pairs by sampling outputs from a single SR model after SFT using varying noise seeds. However, under the rigid spatial constraints of the SR task, these noise-induced variations often degrade into mere random texture hallucination rather than genuine structural diversity. Optimizing over this narrow candidate pool forces the model to fit meaningless artifacts, drastically limiting preference alignment.
To bridge these gaps, we argue that a robust post-training framework for Real-ISR requires two key innovations: \emph{a process-aware, quality-adaptive reward model}, and \emph{an online exploration strategy} that breaks the pseudo-diversity bottleneck.

In this paper, we propose \textbf{OARS} (\textbf{O}nline \textbf{A}lignment for \textbf{R}eal-world I\textbf{S}R), a progressive online RL framework driven by an MLLM-based reward, \textbf{COMPASS} (\textbf{COM}posite \textbf{P}rocess-\textbf{A}ware \textbf{S}R \textbf{S}core).

\textbf{COMPASS} adopts a process-oriented view of enhancement: instead of scoring outputs in isolation, it evaluates the \emph{LR$\rightarrow$SR} transition along two complementary factors, Fidelity for content and structure preservation, and Perceptual Gain for perceptual improvement relative to the input. 
To provide reliable supervision across diverse degradations, we curate \textbf{COMPASS-20K} with synthetic LR inputs paired with ground truth and real-world low-quality (LQ) inputs without references, where each input is associated with multiple SR outputs from diverse enhancement models.
To ensure the perceptual scores are both globally comparable across diverse images and sensitive to subtle differences among SR outputs from the same LR, we further introduce a three-stage perceptual annotation pipeline that combines global anchor scoring, fine-grained intra-group ranking, and rank-guided score calibration. 
Based on COMPASS-20K, we train a unified reward model. 
We then design an input-quality-adaptive scheme that gates the perceptual gain term by fidelity and adjusts its strictness according to the LR quality score, discouraging hallucinations on high-quality inputs while allowing larger improvements for severely degraded ones. Moreover, COMPASS achieves state-of-the-art pairwise preference accuracy on the SR quality assessment benchmark SRIQA-Bench\cite{chen2025toward}.

On top of this reward, \textbf{OARS} performs progressive online alignment in three stages. 
It starts with a cold-start stage that learns basic SR capability via flow-matching SFT on paired LR--HR data. 
It then introduces a full-reference RL stage to stabilize early optimization with GT-based consistency rewards computed directly by distance metrics. 
Finally, it scales to unpaired real-world LQ in a non-reference RL stage guided solely by COMPASS. 
To avoid the exploration collapse of offline sampling from a rigid SR policy, OARS performs shallow policy optimization via LoRA on the base generative model, leveraging its higher stochasticity for on-policy exploration. 
Training is further stabilized by a negative-aware objective with group-wise reward normalization and variance-based filtering, which helps suppress reward hacking while maintaining fidelity. 
In inference, the learned LoRA is merged into the cold-start SR model to obtain the final aligned policy. 

Our main contributions are summarized as follows:
\begin{itemize}
    \item We curate \textbf{COMPASS-20K}, a comprehensive enhancement dataset covering diverse degradations, featuring a novel three-stage annotation pipeline to capture fine-grained intra-group variations.
    
    \item We propose \textbf{COMPASS}, a process-aware MLLM reward with an input-quality-adaptive mechanism to balance fidelity and perceptual gain, outperforming prior IQA-based rewards and achieving state-of-the-art accuracy on the SR preference benchmark
    
    \item We present \textbf{OARS}, an online RL framework that integrates Cold Start SFT, Full-Reference RL, and Non-Reference RL. By using base-model stochasticity to overcome pseudo-diversity and utilizing a negative-aware objective via LoRA shallow fine-tuning, OARS achieves state-of-the-art performance.
\end{itemize}

\section{Related Work}
\noindent\textbf{Generative Real-ISR.} Early deep learning approaches \cite{dong2014learning,lim2017enhanced,zhang2018image,liang2021swinir,chen2023activating,cao2023ntire} primarily rely on regression-based architectures optimized with pixel-wise losses like $\ell_1$, $\ell_2$, or PSNR-oriented objectives. 
GAN-based approaches \cite{wang2021real,zhang2021designing} introduce adversarial learning to enhance realism.
Recently, diffusion and flow-based methods have demonstrated superior generative capacity, and have emerged as the dominant paradigm for Real-ISR. Leveraging large-scale pretrained image generation priors \cite{stabilityai_stablediffusion,blackforestlabs_flux2024}, approaches including StableSR \cite{wang2024exploiting}, DiffBIR \cite{lin2024diffbir}, SeeSR \cite{wu2024seesr}, and FluxSR \cite{li2025one} exploit SFT and distribution matching to synthesize details. These models enjoy higher robustness to real degradations \cite{wu2024one,lin2025harnessing,sun2401improving,yue2025arbitrary,chen2025adversarial,wang2025gendr,chenimproved}.\vspace{2pt}

\noindent\textbf{IQA.}
Traditional IQA can mainly be divided into FR methods, which evaluate degradation by comparing distorted images against pristine references~\cite{wang2004image,sheikh2006image,ding2020image}, and NR approaches, which estimate perceptual quality blindly via learned data priors \cite{kang2014convolutional, ke2021musiq,wang2023exploring}. Recently, MLLMs have advanced this field by generating either numerical ratings \cite{wu2023QAlign,you2025teaching}, or interpretable textual descriptions \cite{you2024DQA,zhang2025qeval}. However, these approaches rely heavily on extensive text annotations for SFT. To alleviate this and improve generalization, recent methods apply RL \cite{li2025qinsight, zhang2025vqinsight,wu2025visualquality}, enabling MLLMs to jointly deliver accurate quality scores and linguistic reasoning. \vspace{2pt}

\noindent\textbf{Preference Alignment.} The alignment of generative models to human preferences has become an important research direction in MLLMs and image generation \cite{dai2023safe,ziegler2019fine,ouyang2022training,wu2025qwen}. RLHF \cite{christiano2017deep} optimizes models using learned reward functions trained on human preference data. DPO \cite{rafailov2023direct,dai2023safe} directly optimizes policies utilizing pairwise preference data under a Bradley–Terry formulation. Diff-DPO \cite{wallace2024diffusion} and subsequent extensions \cite{liang2024step,hu2025towards,liu2025videodpo} adapt DPO to diffusion models by defining preference losses at denoising steps. More recently, online RL \cite{sutton1998reinforcement} has been introduced to generative diffusion and flow-based models. Flow-GRPO \cite{liu2025flow} integrates policy-gradient RL into flow matching via ODE-to-SDE reformulation to enable on-policy explorations, while DiffusionNFT \cite{zheng2025diffusionnft} performs negative-aware online fine-tuning directly on the forward process to improve training effectiveness and stability. These approaches demonstrate the promise of online RL for alignment.


\section{COMPASS: COMposite Process-Aware SR Score}\label{sec:method1}
\begin{figure*}[t]
    \centering
    \includegraphics[width=1\linewidth]{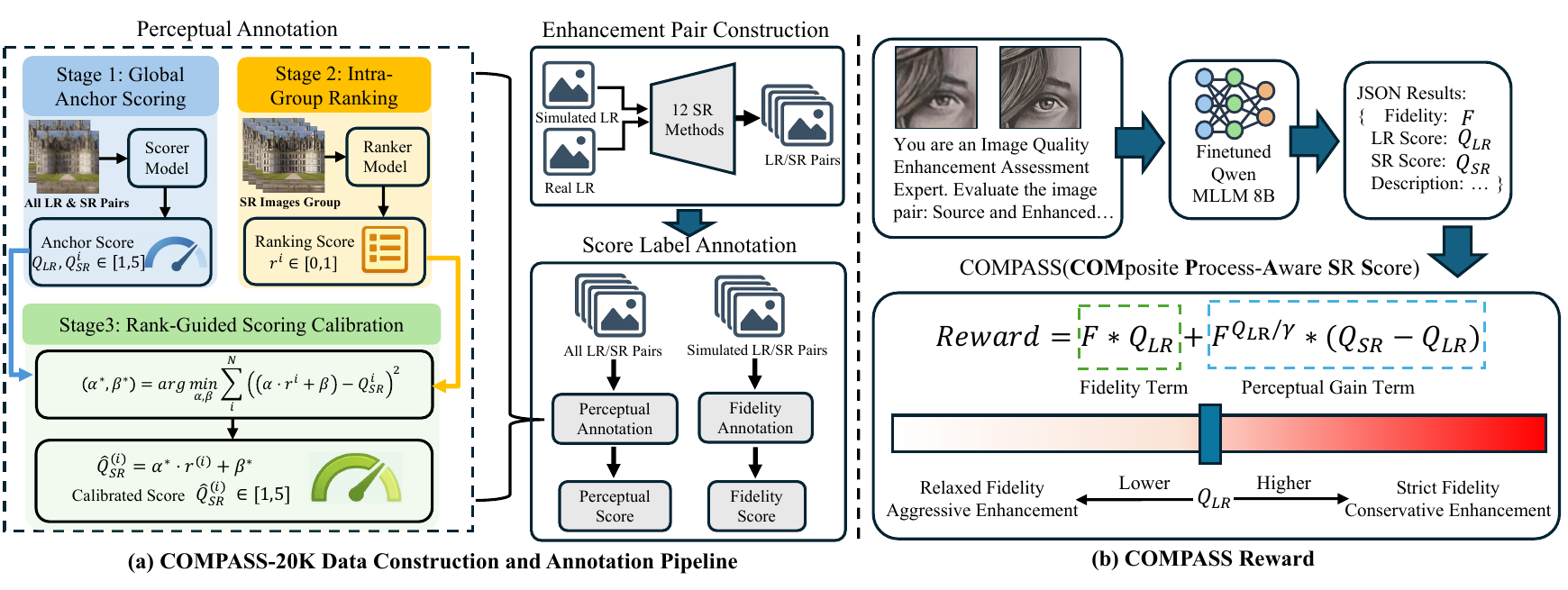}
    \vspace{-20pt}
    \caption{\textbf{Overview of the COMPASS-20K dataset construction and the COMPASS reward framework.} (a) Data generation and score label annotation pipeline. (b) COMPASS uses an input-quality-adaptive mechanism to balance fidelity preservation and perceptual gain conditioned on input quality.}
    \label{fig:compass_pipeline}
    \vspace{-20pt} %
\end{figure*}
To provide a robust alignment signal for our subsequent online reinforcement learning, this section presents the proposed \textbf{COMPASS} reward framework. We detail the construction of the \textbf{COMPASS-20K} dataset and the hierarchical annotation pipeline, culminating in the final MLLM reward model.

\subsection{Motivation}
A practical RL reward for Real-ISR should reflect human preference: the output must look better and remain a plausible transformation of the input. However, existing IQA metrics present two fundamental limitations. First, the metrics themselves are inadequate: Full-Reference (FR) metrics require unavailable GT, while No-Reference (NR) metrics lack the fine-grained sensitivity for generative SR. Second, they are inherently \emph{output-centric}---evaluating the restored image merely as a static outcome and entirely ignoring the starting LR condition.

To overcome these limitations, we must reconstruct the evaluation paradigm starting from the data level. A major bottleneck is the absence of comprehensive datasets featuring real-world degraded images, diverse generative SR outputs, and fine-grained preference annotations. Therefore, we first curate a dedicated enhancement dataset to fill this gap. To resolve the inherent drawbacks of existing FR and NR metrics, we design an annotation protocol that explicitly treats Real-ISR as a \emph{process} ($LR \rightarrow SR$). Rather than scoring a static outcome, we evaluate this transition by decoupling it into \textbf{Fidelity} preservation and \textbf{Perceptual Gain}.

\subsection{COMPASS-20K}
To cover both synthetic and real-world degradation scenarios, we construct a diverse initial set of 2,400 images. Specifically, we apply Real-ESRGAN-style\cite{wang2021real} random degradations to the DIV2K\cite{Agustsson_2017_CVPR_Workshops} dataset to generate 800 synthetic low-resolution (LR) images. Additionally, we collect 1,600 real-world low-quality (LQ) images from the Internet, encompassing a wide range of practical scenes. We then apply 12 mainstream image enhancement algorithms to these 2,400 inputs, resulting in 28,800 LR--SR image pairs. For notational simplicity, we denote both synthetic LR and real LQ inputs as $x_{\mathrm{LR}}$.

For each $LR \rightarrow SR$ enhancement process, we annotate two dimensions---\textbf{Fidelity} and \textbf{Perceptual Gain}---and further provide textual descriptions for interpretability. Perceptual quality is annotated for all data, while fidelity is annotated only on the DIV2K subset, since real-world degraded images do not have ground-truth references.

\textbf{Fidelity Annotation.}
Fidelity measures content consistency in the enhancement process. Following common practice, we use DISTS\cite{ding2021locally} to capture structural and textural similarity, and on the synthetic subset, we define fidelity using the SR--GT DISTS distance. For reward modeling, we normalize it to $[0,1]$ as
\begin{equation}
F = 1 - \mathrm{DISTS}(x_{\mathrm{SR}}, x_{\mathrm{GT}}),
\end{equation}
where $F=1$ indicates the best fidelity and $F=0$ the worst.

\textbf{Perceptual Quality-Gain Annotation.}
A naive solution is to score LR and SR independently with an NR-IQA model and take the difference, but this is suboptimal for enhancement. Existing NR-IQA models are not optimized for modern enhancement outputs, especially generative ones. In addition, differences among SR results from the same LR input are often subtle, and absolute cross-image scoring lacks sufficient sensitivity for such intra-group distinctions. Direct MOS labeling is also expensive and less reliable for fine-grained comparisons.

While the pioneering work DiffIQA \cite{chen2025toward} provides valuable fine-grained pairwise evaluations, it contains only partial intra-group comparisons that cannot yield a complete ranking score. To address this, we propose a \textbf{three-stage alignment annotation pipeline} that combines global comparability with comprehensive intra-group discriminability.

\textbf{Stage 1: Global Anchor Scoring.}
We use Q-Insight \cite{li2025qinsight} score model to score both LR and SR images, obtaining $Q_{\mathrm{LR}}$ and the anchor SR score $Q_{\mathrm{SR}}$, both in $[1,5]$. These scores provide a globally comparable quality scale.

\textbf{Stage 2: Intra-Group Ranking.}
To capture the subtle quality differences among SR results derived from the same LR input, we establish a fine-grained intra-group ranking. We first build a dedicated pairwise comparison model upon the Q-Insight compare model and train it using the DiffIQA dataset as supervision. We then deploy this model to conduct exhaustive pairwise comparisons among all SR outputs within each group. Finally, we aggregate these pairwise outcomes into a continuous relative ranking score $r\in[0,1]$, where 1 and 0 denote the best and worst SR results within that specific group, respectively.

\textbf{Stage 3: Rank-Guided Scoring Calibration.}
To preserve the intra-group ranking while aligning it with the global quality scale, we perform a per-group linear calibration. For a group with $N$ SR results, we estimate a scale $\alpha$ and bias $\beta$ by
\begin{equation}
(\alpha^*, \beta^*)=\arg\min_{\alpha,\beta}\sum_{i=1}^{N}\left(\alpha \cdot r^{(i)}+\beta-Q_{\mathrm{SR}}^{(i)}\right)^2.
\end{equation}
The calibrated SR quality score is then defined as
\begin{equation}
\hat{Q}_{\mathrm{SR}}^{(i)}=\alpha^* \cdot r^{(i)}+\beta^*.
\end{equation}

This score preserves the fine-grained intra-group ranking while remaining aligned with the global scale, making it suitable for reward supervision. $\hat{Q}_{\mathrm{SR}}^{(i)}$ denotes the i-th calibrated SR quality score in the group.

\textbf{Interpretability Annotation.}
We additionally generate concise, process-aware textual rationales for each LR--SR pair using Qwen3-VL-32B~\cite{Qwen3-VL}, describing fidelity-related changes and perceptual-gain effects. 
Importantly, these rationales are used for data quality control: we manually check the top/bottom 5\% scored samples and discard pairs with obvious score--description conflicts, reducing corner cases and improving annotation reliability.

\noindent Detailed dataset construction procedures are provided in the \textbf{supplementary materials}.

\subsection{COMPASS Reward}
After constructing the COMPASS-20K dataset, we perform full-parameter SFT on QwenVL-8B\cite{Qwen3-VL}. For samples with fidelity annotations, the model is trained to jointly predict fidelity and perceptual quality scores; for samples without fidelity annotations, only quality prediction is supervised. We use prompts to indicate the data type and task setting, allowing a unified model to handle different annotation configurations.

At inference time, the model generates both numerical scores and a textual description. The numerical outputs consist of the predicted fidelity score $F$ for the enhancement process, the perceptual quality score of the input image $Q_{\mathrm{LR}}$, and the perceptual quality score of the enhanced image $Q_{\mathrm{SR}}$, where $F \in [0, 1]$ and $Q_{\mathrm{LR}}, Q_{\mathrm{SR}} \in [1, 5]$.
To unify fidelity and perceptual quality change into a single reward signal, we propose an input-quality-adaptive mechanism that balances content preservation and quality improvement according to the input quality. Denoting the perceptual gain introduced by enhancement as $\Delta Q = Q_{\mathrm{SR}} - Q_{\mathrm{LR}}$, we define the reward as
\begin{equation}
R = F\cdot Q_{\mathrm{LR}} + F^{Q_{\mathrm{LR}}/\gamma}\cdot \Delta Q,
\label{reward}
\end{equation}
where $\gamma$ is a hyper-parameter. We set $\gamma=7$ based on ablation studies.

The first term, $F\cdot Q_{\mathrm{LR}}$, measures how much of the input image's original quality is preserved. The second term, $F^{Q_{\mathrm{LR}}/\gamma}\cdot \Delta Q$, captures the additional perceptual gain. Crucially, this gain is subjected to an input-quality-adaptive control via the exponent $Q_{\mathrm{LR}}/\gamma$. When the input quality is high, this exponent becomes larger, making the reward highly sensitive to fidelity drops and thus encouraging conservative enhancement. Conversely, for low-quality inputs, this fidelity constraint relaxes, allowing for more aggressive perceptual improvements. Ultimately, this dynamic gating mechanism ensures that any perceptual enhancement is strictly bounded by content preservation, effectively balancing the perception-fidelity trade-off according to the initial degradation severity.

\section{OARS: Online Alignment for Real-world ISR}
\begin{figure}[!t]
\centering
\includegraphics[width=\linewidth]{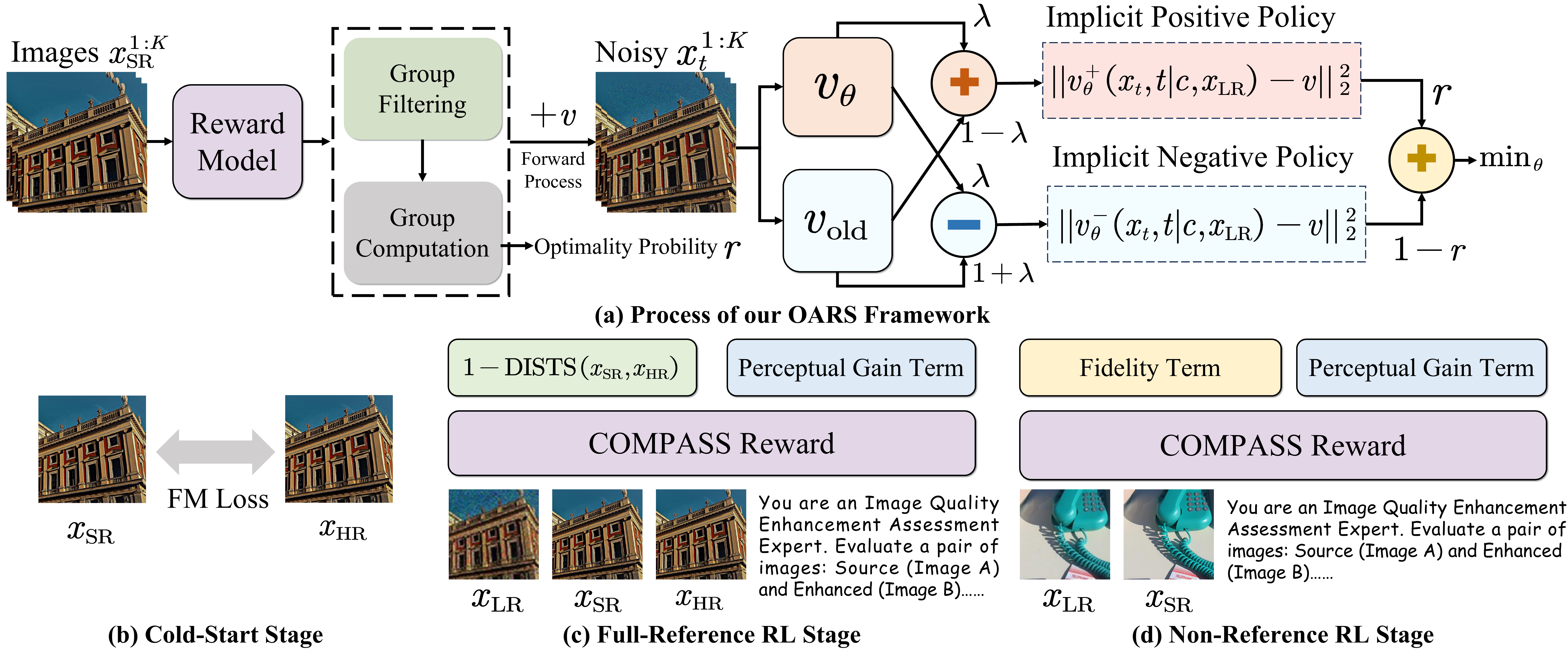}
\vspace{-20pt}
\caption{
Overview of the proposed OARS framework.
(a) Policy optimization process of our OARS framework;
(b) Cold-start stage with paired LR--HR supervision;
(c) Full-reference RL stage;
(d) Non-reference RL stage on unpaired data guided by our reward.
}
\label{fig:framework}
\vspace{-15pt}
\end{figure}

After obtaining the COMPASS reward that reflects both fidelity and perceptual quality, we propose a progressive forward-process online RL framework. Through multi-stage continual learning, it can transform a general generative large model into an SR expert that aligns with human preferences. As shown in Fig.~\ref{fig:framework}, our framework includes three stages: cold start, full-reference reinforcement learning, and non-reference reinforcement learning. 

\textbf{Cold Start Stage.}
To endow the model with initial super-resolution capability, we first warm up the model on large-scale LR--HR image pairs using a Flow Matching objective. This stage serves as a cold start that equips the model with basic perceptual reconstruction ability before reinforcement fine-tuning.

Let $x_{\mathrm{HR}}$, $x_{\mathrm{LR}}$, and $c$ respectively denote high-resolution input, its corresponding low-resolution input, and the enhanced prompt. Given a Gaussian noised sample $\epsilon_{\mathrm{noise}} \sim \mathcal{N}(0, \mathbf{1})$, rectified flow defines an interpolated noisy sample at time $t \in [0,1]$ as:
$x_t = (1 - t)\, x_{\mathrm{HR}} + t\, \epsilon_\mathrm{noise}.$ Then, we train a conditional velocity field network $v_\theta(x_t, t \mid c, x_{\mathrm{LR}})$ to approximate the target velocity
$v = \epsilon_{\mathrm{noise}} - x_{\mathrm{HR}}$
by minimizing the following loss:
\begin{equation}
\mathcal{L}_{\mathrm{SFT}}(\theta)
=
\mathbb{E}_{\substack{
t \sim \mathcal{U}(0,1),
x_{\mathrm{HR}},
\epsilon_{\mathrm{noise}}
}}
\left[
\left\|
v - v_\theta(x_t, t \mid x_{\mathrm{LR}}, c)
\right\|_2^2
\right].
\end{equation}
Through this cold start stage, the model learns a continuous conditional transport dynamics that maps samples from the noise distribution to the high-resolution image distribution conditioned on low-resolution inputs. It provides a strong initialization with reliable perceptual reconstruction capability.

\textbf{Full-Reference RL Stage.}
After obtaining the SFT model $v_{\theta_{\mathrm{SFT}}}$ from the cold start stage, a straightforward approach is to further perform RL on top of the SFT weights using additional low-resolution data. However, we observe two critical issues with this strategy: \textbf{1)} Directly optimizing both perceptual quality and structural consistency on out-of-domain data is highly challenging, leading to unstable training and poor convergence; \textbf{2)} RL initialized from $v_{\theta_{\mathrm{SFT}}}$ tends to exploit the reward signal by over-optimizing perceptual scores while ignoring content and structural fidelity, resulting in severe reward hacking.

To address these issues, we introduce a reference-based RL stage as a buffer between SFT and reference-free optimization. In this stage, RL is still performed on LR inputs with available HR references. To reduce optimization difficulty and improve the reliability of consistency supervision, we directly compute the consistency reward using distance-based metrics (e.g., $\mathrm{DISTS}$) between the generated image $x_{\mathrm{SR}}$ and the ground-truth image $x_{\mathrm{HR}}$, instead of relying on the learned reward model to predict consistency shown in Eq.~\ref{reward}.

Inspired by shallow fine-tuning~\cite{wu2025shadow, liu2024tuning, fleshman2024re}, we do not apply RL directly on the SFT weights. Instead, we perform policy optimization by applying LoRA-based updates on the base model weights $v_{\theta_{\mathrm{base}}}$. During inference, the learned LoRA parameters are merged into the SFT model to obtain the final policy. This design is motivated by the following observations: 1) The parameter distributions of $v_{\theta_{\mathrm{base}}}$ and $v_{\theta_{\mathrm{SFT}}}$ are closely aligned, enabling stable merging without introducing abnormal outputs; 2) Sampling rollouts from the base model exhibits higher stochasticity, which is beneficial for exploration during reinforcement learning; 3) Optimizing on the base model is less prone to reward hacking and better preserves fidelity and structural consistency.

Since SR can be treated as a strongly constrained generation task that does not heavily rely on step-wise trajectory optimization~\cite{liu2025flow}, and considering the optimization efficiency, we adopt a forward-process RL paradigm~\cite{zheng2025diffusionnft}. Given $K$ samples $x^{1:K}_{\mathrm{SR}}$ from the same LR input and enhanced prompt, we first evaluate each sample via our reward to obtain raw reward scores $s^{1:K}_{\mathrm{raw}}$. 
Notably, the fidelity term in this stage is directly computed between the $x_{\mathrm{HR}}$ and the $x_{\mathrm{SR}}$, rather than being predicted by our reward. To avoid ambiguous supervision caused by nearly identical samples, we discard groups whose raw rewards exhibit a high mean but low variance, as such samples provide limited discriminative learning signals. 
After filtering, we perform group computation on the remaining raw rewards, followed by clipping, to obtain the final optimal probability $r \in [0,1]$. Formally, the optimal probability is computed as: $r = 0.5 + 0.5\,\operatorname{clip}\!\left(\left({s_{\mathrm{raw}} - \mathbb{E}_{\pi^{\mathrm{old}}(\cdot \mid c)}[s_{\mathrm{raw}}}]\right)/\text{std}, -1, 1 \right)$, where $\text{std} > 0$ is the standard deviation of raw rewards in the group and $\pi^{\mathrm{old}}$ denotes the reference policy.

Furthermore, we define a contrastive objective that steers the velocity predictor toward high-reward behaviors while suppressing low-reward ones. Formally, let $v$ denote the target velocity field, and let $v_\mathrm{old}$ be the frozen reference policy. We define implicit positive and negative policies as linear combinations of the old policy and the current policy $v_\theta$:
\begin{align}
v_\theta^{+}(x_t, t \mid c, x_{\mathrm{LR}}) &= (1 - \lambda)\, v_{\mathrm{old}}(x_t, t \mid c, x_{\mathrm{LR}}) + \lambda\, v_\theta(x_t, t \mid c, x_{\mathrm{LR}}), \\
v_\theta^{-}(x_t, t \mid c, x_{\mathrm{LR}}) &= (1 + \lambda)\, v_{\mathrm{old}}(x_t, t \mid c, x_{\mathrm{LR}}) - \lambda\, v_\theta(x_t, t \mid c, x_{\mathrm{LR}}),
\end{align}
where $\lambda$=1.0. The final RL objective is defined as: 
\begin{equation}
\mathcal{L}_{\mathrm{RL}}(\theta)
=
\mathbb{E}
\Big[
r \, \big\| v_\theta^{+}(x_t, t \mid c, x_{\mathrm{LR}}) - v \big\|_2^2
+
(1 - r)\, \big\| v_\theta^{-}(x_t, t \mid c, x_{\mathrm{LR}}) - v \big\|_2^2
\Big].
\label{nft}
\end{equation}
By explicitly modeling the positive and negative policy directions, this negative-aware objective effectively balances perceptual quality enhancement and reference-based consistency preservation, getting the optimized lora weights $\Delta_{\mathrm{FR}}$.

\textbf{Non-Reference RL Stage.}
In the final stage, we aim to further improve the perceptual quality while ensuring that fidelity degradation is minimized as much as possible. To this end, we perform RL on additional out-of-domain low-resolution data without ground-truth references in our COMPASS-20K, enabling the model to generalize beyond the training distribution. Building upon the learned LoRA $\Delta_{\mathrm{FR}}$, we continue training the model via LoRA finetuning. In the non- reference setting, no ground-truth high-resolution images are available. Therefore, the reward signal is entirely provided by the COMPASS reward model introduced in Eq.~\ref{reward}. Noted that, we still perform RL on the base model $v_{\theta_{\mathrm{base}}}$. The policy optimization objective follows the same process as in the reference-based stage (Eq.~\ref{nft}). Thus, we can get the final lora weights $\Delta_{\mathrm{NR}}$. \textbf{During inference}, we merge the learned LoRA parameters $\Delta_{\mathrm{NR}}$ from the final stage into the backbone weights obtained from the cold start (SFT) stage $v_{\theta_{\mathrm{SFT}}}$.


\section{Experiment}
\subsection{Experimental Setup}
\textbf{Reward Implementation and Evaluation.} We train the reward on our COMPASS-20K, whose data generation and annotation procedure are described in Sec.~\ref{sec:method1}, using 19,200 annotated pairs in total, with half from real-world LQ inputs and half from synthetic LR inputs. We adopt Qwen-3-VL-8B-Instruct\cite{Qwen3-VL} as the initialization and perform full-parameter supervised fine-tuning. Training is conducted on 8 NVIDIA H20 GPUs (96GB) with a global batch size of 64 for 8 epochs. We use AdamW for optimization, with an initial learning rate of 1e-5 and a cosine decay schedule.

To verify whether the learned reward aligns with human preference, we evaluate it on SRIQA-Bench\cite{chen2025toward}, which contains 100 LR images and results produced by multiple SR methods. The benchmark provides pairwise comparisons between SR outputs for the same LR input, together with ground-truth preference annotations. Given a pair of SR candidates $(x_{\mathrm{SR}}^a, x_{\mathrm{SR}}^b)$, we compute their reward scores and predict the preferred one by comparing the two scalar rewards. We then report \textbf{pairwise preference accuracy}, i.e., the percentage of comparisons where the reward-based ordering matches the annotated human preference. This evaluation directly measures the reward model’s capability of ranking competing SR outputs in a fine-grained and perceptually consistent manner.

\noindent\textbf{Online RL Implementation and Evaluation.} We adopt Qwen-Image-Edit-2509~\cite{wu2025qwen} as the base model and perform LoRA fine-tuning with rank=$32$ and alpha=$64$. The per-GPU batch size is set to $1$, and the model is optimized using Adam with a learning rate of $3\times10^{-4}$ and weight decay $1\times10^{-4}$. During training, the diffusion model uses $6$ sampling steps for training and $40$ steps for inference. For group filtering, the thresholds on the reward mean and variance are set to $0.9$ and $0.05$, respectively. For each $x_{\mathrm{LR}}$, we sample $K=24$ candidate groups. All RL experiments are conducted on a single computer with $8$ NVIDIA H20 GPUs ($96$\,GB). The reward server is deployed on a computer with $8$ NVIDIA A100 GPUs ($80$\,GB) using vLLM for inference. The cold-start and full-reference RL stages are trained on the LSDIR dataset~\cite{LSDIR}, while the non-reference RL stage is trained on our constructed COMPASS-20K, using a split that is strictly separated from the reward training set and contains 800 real-world LQ inputs. We evaluate the SR performance of our model on RealSR~\cite{cai2019toward}, DIV2K~\cite{agustsson2017ntire}, and RealSet80~\cite{yue2024efficient}, with $512 \times 512$ resolution, performing $4\times$ SR.
FR metrics include PSNR~\cite{wang2004image}, SSIM~\cite{wang2004image}, LPIPS, and DISTS, 
while NR metrics include LIQE~\cite{zhang2023blind}, MUSIQ~\cite{ke2021musiq}, MANIQA~\cite{yang2022maniqa}, Q-Insight~\cite{li2025qinsight}, and TOPIQ~\cite{chen2024topiq}.
\begin{table*}[t]
\centering
\caption{\textbf{Comparison of different IQA methods} on SRIQA-Bench \cite{chen2025toward}. Throughout this paper, the best, second-best, and third-best results are highlighted in \best{bold red}, \secondbest{underlined blue}, and \third{italic green}.}
\vspace{-10pt}

\setlength{\tabcolsep}{6pt}
\renewcommand{\arraystretch}{1.15}
\small
\resizebox{0.7\textwidth}{!}{
\begin{tabular}{l | l | c c c}
\shline
Type & Method & Reg-Acc & Gen-Acc  & All-Acc \\
\hline \hline

\multirow{7}{*}{\begin{tabular}[c]{@{}l@{}}GT-Ref \\\end{tabular}}
& PSNR   & 80.7 & 41.7 & 34.7 \\
& SSIM\cite{wang2004image}   & 83.0 & 45.3 & 37.4 \\
& LPIPS\cite{zhang2018unreasonable}  & 84.7 & 63.8 & 72.2 \\
& DISTS\cite{ding2020image}  & \secondbest{86.0} & 63.9 & 71.4 \\
& AHIQ\cite{9857006}   & 71.0 & 71.5 & 69.6 \\
& TOPIQ\cite{chen2024topiq}  & 78.3 & \third{73.0} & 77.7 \\
& A-FINE\cite{chen2025toward} & 83.3 & \best{78.9} & \secondbest{82.4} \\
\hline

\multirow{4}{*}{\begin{tabular}[c]{@{}l@{}}GT-Free \\ \end{tabular}}
& CKDN\cite{zheng2021learning}      & 76.7  & 64.3 & 59.1 \\
& CLIP-IQA+\cite{wang2023exploring} & 80.3 & 72.5 & \third{79.5} \\
& Q-Insight\cite{li2025qinsight} & \best{88.0} & 71.0 & 78.8 \\
& COMPASS (Ours)      & \third{85.0} & \secondbest{78.3} & \best{83.1} \\
\shline
\end{tabular}
}
\label{tab:iqa_compare}
\vspace{-15pt}
\end{table*}
\subsection{Reward Experimental Results}
To validate the effectiveness of our reward model, we compare it against two categories of visual quality assessment methods: (1) \textbf{GT-Ref} metrics (e.g., PSNR, LPIPS, A-FINE), which rely on GT references to compute quality scores; and (2) \textbf{GT-Free} metrics, which operate without GT. Within the GT-Free category, CKDN and our method use the LR input as a conditional reference, whereas Q-Insight and CLIP-IQA+ are reference-free. As shown in Table~\ref{tab:iqa_compare}, our method achieves the highest overall accuracy (\textbf{83.1\%}) across all 5500 test samples, outperforming all GT-Ref and GT-Free baselines. Beyond improved alignment with human perception, our approach is GT-Free and thus intrinsically suited for Real-ISR where HR references are unavailable. 

\begin{table*}[t]
\centering
\caption{Quantitative comparison on three classical SR datasets, including RealSR, DIV2K, and RealSet80. $\uparrow/\downarrow$ indicates higher/lower is better.}
\vspace{-6pt}
\renewcommand{\arraystretch}{1.15}
\resizebox{\linewidth}{!}{
\begin{tabular}{c | l | c c c c | c c c c c}
\hline
Dataset & Method
& PSNR$\uparrow$ & SSIM$\uparrow$ & LPIPS$\downarrow$ & DISTS$\downarrow$
& LIQE$\uparrow$ & MUSIQ$\uparrow$ & MANIQA$\uparrow$ & Q-Insight$\uparrow$ & TOPIQ$\uparrow$ \\
\hline\hline

\multirow{9}{*}{\textbf{RealSR}}
& DiffBIR~\cite{lin2024diffbir}  & 23.20 & 0.6346 & 0.3350 & 0.2162 & 3.5529 & 65.25 & 0.4620 & 3.530 & 0.6033 \\
& OSEDiff~\cite{wu2024one}  & 23.07 & \third{0.6850} & \third{0.2941} & \secondbest{0.2109} & \secondbest{4.0681} & \third{68.95} & 0.4876 & \best{3.712} & \third{0.6441} \\
& SeeSR~\cite{wu2024seesr}    & \best{24.34} & \best{0.7187} & \best{0.2754} & \third{0.2134} & 3.3938 & 65.53 & 0.4856 & 3.285 & 0.6246 \\
& SinSR~\cite{wang2024sinsr}   & \third{23.68} & 0.6649 & 0.3490 & 0.2445 & 3.2255 & 61.03 & 0.4230 & 3.264 & 0.5383 \\
& StableSR~\cite{wang2024exploiting} & \secondbest{23.73} & \secondbest{0.6979} & \secondbest{0.2792} & \best{0.2023} & 3.0532 & 61.65 & 0.3826 & 3.480 & 0.5201 \\
& PURE~\cite{wei2025perceive}     & 21.31 & 0.5738 & 0.3859 & 0.2468 & 3.7881 & 66.57 & 0.4829 & 3.569 & 0.6301 \\
& UARE~\cite{li2025uare}     & 21.38 & 0.6464 & 0.3095 & 0.2344 & \third{4.0658} & \secondbest{69.67} & \secondbest{0.5260} & \third{3.664} & \secondbest{0.6796} \\
& Qwen-SFT~\cite{wu2025qwen} & 22.71 & 0.6462 & 0.3100 & 0.2203 & 3.8146 & 68.57 & \third{0.4897} & 3.545 & 0.6397 \\
& OARS (Ours)     & 22.36 & 0.6481 & 0.3095 & 0.2244 & \best{4.3045} & \best{71.41} & \best{0.5277} & \secondbest{3.701} & \best{0.6803} \\

\hline\hline

\multirow{9}{*}{\textbf{DIV2K}}
& DiffBIR~\cite{lin2024diffbir}  & \third{18.94} & 0.4332 & 0.4009 & 0.2238 & 3.8573 & 67.20 & 0.4574 & 3.577 & 0.6467 \\
& OSEDiff~\cite{wu2024one}  & 18.86 & \third{0.4563} & \best{0.3579} & 0.2209 & 3.8877 & 67.83 & 0.4422 & 3.722 & 0.6269 \\
& SeeSR~\cite{wu2024seesr}   & \secondbest{19.11} & \secondbest{0.4580} & \secondbest{0.3769} & 0.2339 & 3.7445 & 66.31 & 0.4686 & 3.611 & 0.6330 \\
& SinSR~\cite{wang2024sinsr}    & 18.58 & 0.4059 & 0.4483 & 0.2455 & 3.4629 & 64.12 & 0.4483 & 3.224 & 0.5997 \\
& StableSR~\cite{wang2024exploiting} & \best{19.85} & \best{0.4940} & 0.4796 & 0.2887 & 1.8466 & 43.25 & 0.2181 & 2.066 & 0.3276 \\
& PURE~\cite{wei2025perceive}     & 16.71 & 0.3661 & 0.4449 & 0.2293 & \third{4.2701} & 70.06 & \best{0.5201} & 3.721 & 0.6621 \\
& UARE~\cite{li2025uare}     & 16.59 & 0.3857 & 0.4074 & \third{0.2138} & 4.2627 & \third{70.45} & \third{0.5028} & \secondbest{3.823} & \secondbest{0.6864} \\
& Qwen-SFT~\cite{wu2025qwen} & 17.31 & 0.3988 & \third{0.3811} & \best{0.1993} & \secondbest{4.3404} & \secondbest{72.35} & 0.4875 & \third{3.818} & \third{0.6745} \\
& OARS (Ours)     & 17.32 & 0.4072 & 0.3812 & \secondbest{0.2036} & \best{4.6668} & \best{74.07} & \secondbest{0.5052} & \best{3.940} & \best{0.6960} \\

\hline\hline

\multirow{9}{*}{\textbf{RealSet80}}
& DiffBIR~\cite{lin2024diffbir}  & \multicolumn{4}{c|}{} & 4.1113 & 68.10 & \best{0.5527} & 3.684 & \third{0.6736} \\
& OSEDiff~\cite{wu2024one}  & \multicolumn{4}{c|}{} & 4.2251 & 68.88 & 0.4995 & 3.725 & 0.6062 \\
& SeeSR~\cite{wu2024seesr}    & \multicolumn{4}{c|}{} & \secondbest{4.3317} & 69.70 & 0.5362 & \secondbest{3.774} & \secondbest{0.6887} \\
& SinSR~\cite{wang2024sinsr}   & \multicolumn{4}{c|}{} & 3.6613 & 62.78 & 0.4483 & 3.382 & 0.5854 \\
& StableSR~\cite{wang2024exploiting} & \multicolumn{4}{c|}{\Large N/A} & 3.9074 & 67.67 & 0.4682 & 3.562 & 0.6440 \\
& PURE~\cite{wei2025perceive}     & \multicolumn{4}{c|}{} & \third{4.2528} & 69.55 & 0.5215 & \third{3.744} & 0.6647 \\
& UARE~\cite{li2025uare}     & \multicolumn{4}{c|}{} & 4.1804 & \secondbest{70.05} & \third{0.5363} & 3.508 & 0.6446 \\
& Qwen-SFT~\cite{wu2025qwen} & \multicolumn{4}{c|}{}
& 4.1602 
& \third{69.79}
& 0.5171 
& 3.701 
& 0.6539 \\
& OARS (Ours)     & \multicolumn{4}{c|}{} & \best{4.5465} & \best{72.96} & \secondbest{0.5364} & \best{3.823} & \best{0.6967} \\

\hline
\end{tabular}}
\label{tab:maintable}
\vspace{-15pt}
\end{table*}


\subsection{Online RL on Super-Resolution Experimental Results}

\textbf{Quantitative Comparison.} Following prior works~\cite{wu2024one, li2025uare}, we compare our OARS against diffusion-based methods:  StableSR~\cite{wang2024exploiting}, DiffBIR~\cite{lin2024diffbir}, SeeSR~\cite{wu2024seesr}, SinSR~\cite{wang2024sinsr}, OSEDiff~\cite{wu2024one}, Qwen-SFT, the autoregressive method PURE~\cite{wei2025perceive}, and the unified method UARE~\cite{li2025uare}. Note that Qwen-SFT denotes the Qwen-Image-Edit model after the cold-start stage. As reported on Tab.~\ref{tab:maintable}, our OARS achieves consistently top-ranked performance on NR metrics across RealSR, DIV2K, and RealSet80. Moreover, compared to perception-oriented methods such as PURE and UARE, our approach also attains significantly better FR metrics, showing that our reward design and RL strategy effectively balance fidelity preservation and perceptual enhancement. Notably, when compared to Qwen-SFT (our initial policy model), OARS yields consistent and pronounced improvements on all NR metrics, while the FR performance does not exhibit noticeable degradation.

\begin{wrapfigure}{r}{0.4\linewidth}
    \centering
    \vspace{-25pt} 
    \includegraphics[width=1\linewidth]{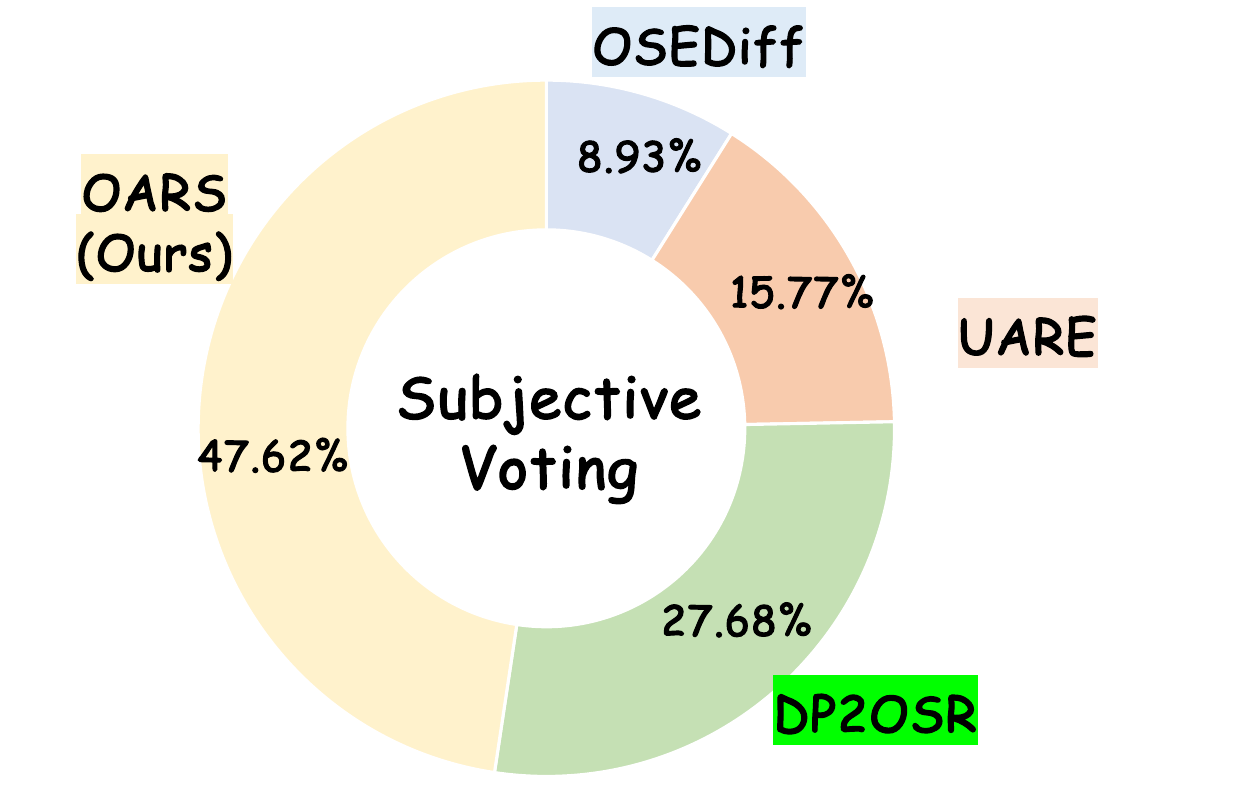} 
    \vspace{-10pt}
    \caption{\textbf{Subjective user study results.} OARS achieves the highest preference rate, receiving 47.62\% of the total votes.}
    \label{fig:user_study}
    \vspace{-10pt}
\end{wrapfigure}
\begin{figure*}[t]
    \centering
    \includegraphics[width=1\linewidth]{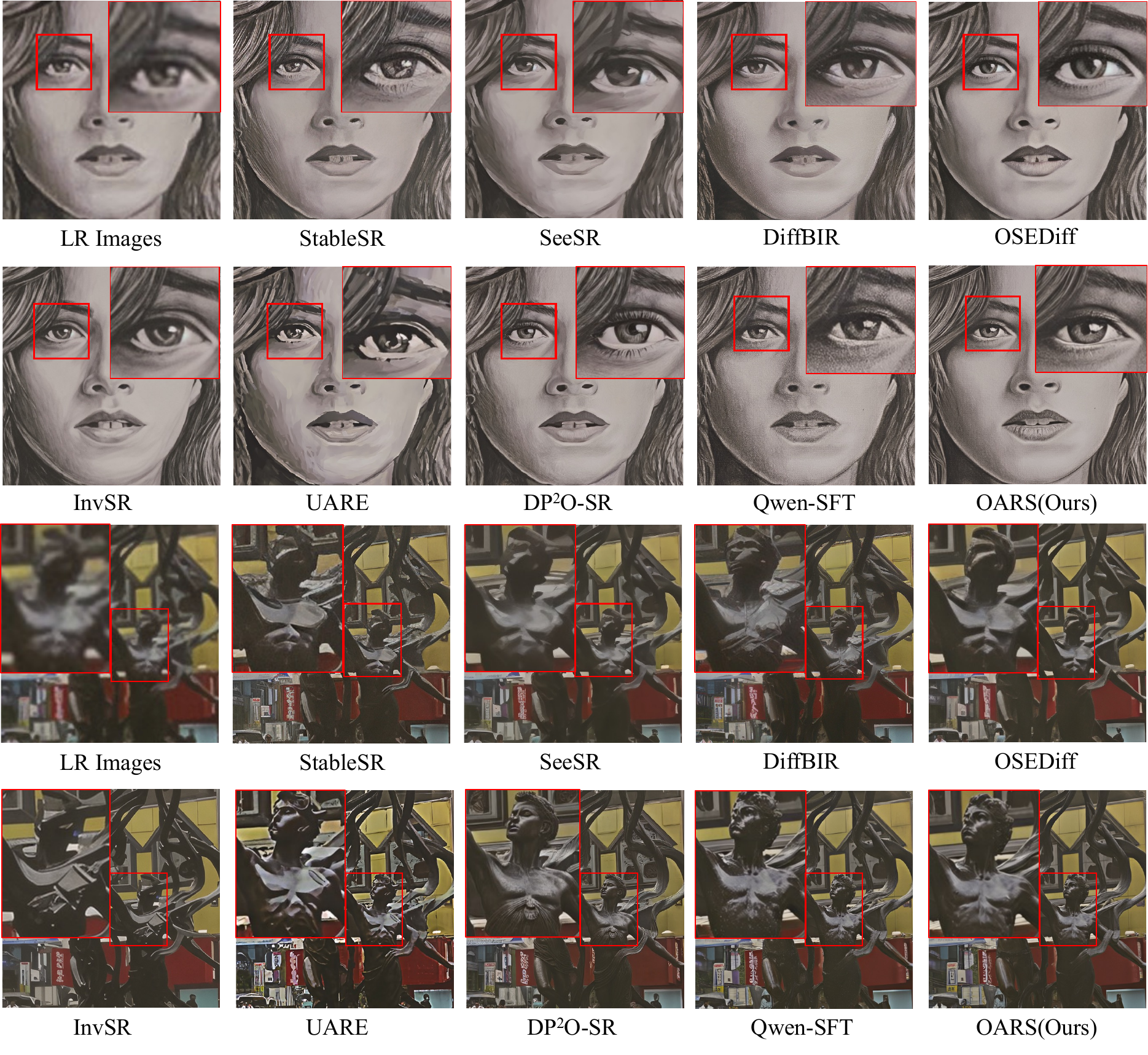}
    \vspace{-20pt}
    \caption{\textbf{Qualitative comparison of OARS against state-of-the-art Real-ISR methods under complex degradations.}}
    \label{fig:subjective_comparison}
    \vspace{-15pt}
\end{figure*}

\noindent\textbf{Qualitative Comparison and User Study.}
Fig.~\ref{fig:subjective_comparison} shows that OARS produces more natural restorations under complex real degradations, recovering fine details while preserving global structure. In comparison, SeeSR, UARE, DP$^2$O-SR, and OSEDiff more frequently exhibit over-smoothing or localized hallucinations/over-sharpening. We further conduct a user study on 12 LR inputs sampled from RealSR, DRealSR, and RealSet80, where 27 expert researchers select the most preferred restored result among the compared methods. OARS receives \textbf{47.62\%} of the votes, higher than the strongest baseline (DP$^2$O-SR, 27.68\%).

\noindent For more quantitative and qualitative comparison results, as well as the detailed user study setup, please refer to the \textbf{supplementary materials}.


\subsection{Ablation Studies}
\textbf{Ablation on Reward Model.} Table~\ref{tab:ablation_gamma} evaluates key components of our reward formulation on COMPASS-20K. Introducing three-stage score calibration (Case~2) improves accuracy over naive absolute scoring (Case~1) by better aligning fine-grained intra-group rankings. Adding explicit fidelity modeling further boosts performance (Case~3), indicating that enforcing structural consistency is crucial. Finally, replacing a fixed linear combination with our Quality-Adaptive $\gamma$ yields the best result, and we select $\gamma=7$ based on the ablation (Case~4).

\noindent\textbf{Ablation on Online RL.}
Tab.~\ref{tab:ablation_rl_stage} analyzes the impact of RL stages and initialization. Starting from Qwen-SFT, applying our stage-1 RL on the \textbf{base} model (Case~1) yields clear perceptual gains across NR metrics while keeping FR metrics stable. Adding stage-2 reference-free RL (Case~2) further improves perceptual quality, giving the best overall NR performance. In contrast, performing RL on the \textbf{SFT} model (Cases~3--4) consistently degrades FR metrics, indicating higher susceptibility to reward hacking and instability. These results validate our progressive RL design and the choice of shallow LoRA optimization on the base model for more effective on-policy exploration and robust alignment.

\noindent\textbf{Ablation on Reward for RL.} We further investigate how different reward signals influence the RL policy. Directly optimizing with naive no-reference metrics induces severe reward hacking, which clearly degrades fidelity and introduces over-generation artifacts. Additionally, dropping either the input-quality-adaptive gating or the fidelity term from our reward formulation results in a severe imbalance between objective fidelity and subjective visual quality. Conversely, employing the complete COMPASS reward achieves the optimal. For more experimental results, please refer to the \textbf{supplementary material}.

\begin{table}[t]
\centering
\caption{\textbf{Ablation study} of dataset calibration, fidelity modeling, and adaptive $\gamma$.}\vspace{-10pt}

\setlength{\tabcolsep}{7pt}
\renewcommand{\arraystretch}{1.15}
\small

\resizebox{0.8\linewidth}{!}{
\begin{tabular}{c | c c c | c}
\shline
\textbf{Case} & \textbf{Score Calibration} & \textbf{Explicit Fidelity} & \textbf{Quality-Adaptive $\gamma$} & \textbf{Accuracy} \\
\hline \hline
1 & \xmark & \xmark & \xmark & 78.8 \\
2 & \cmark & \xmark & \xmark & 81.5 \\
3 & \cmark & \cmark & \xmark & 82.3 \\
4 & \cmark & \cmark & 5   & 82.7 \\
5 & \cmark & \cmark & 7   & \best{83.1} \\
6 & \cmark & \cmark & 9   & 82.8 \\
\shline
\end{tabular}
}

\label{tab:ablation_gamma}
\vspace{-8pt}
\end{table}


\begin{table}[t]
\centering
\caption{\textbf{Quantitative comparison between our OARS and DPO-based SR methods} on RealSet80. Performance gains are reported in parentheses.}
\vspace{-10pt}
\resizebox{0.8\linewidth}{!}{
\renewcommand{\arraystretch}{1.15}
\begin{tabular}{l | c c c c c}
\shline
Method
& LIQE$\uparrow$
& MUSIQ$\uparrow$
& MANIQA$\uparrow$
& Q-Insight$\uparrow$
& TOPIQ$\uparrow$ \\
\hline \hline

Qwen-SFT 
& 4.1602 
& 69.7866 
& 0.5171 
& 3.701 
& 0.6539 \\

+COMPASS   
& 4.5465 \best{{\small(+0.39)}} 
& 72.9598 \best{{\small(+3.17)}} 
& 0.5364 \best{{\small(+0.02)}}  
& 3.823 \best{{\small(+0.12)}} 
& 0.6967 \best{{\small(+0.04)}} \\

+COMPASS+TTT 
& 4.5792 \best{{\small(+0.03)}} 
& 73.2551 \best{{\small(+0.30)}} 
& 0.5428 \best{{\small(+0.01)}}  
& 3.841 \best{{\small(+0.02)}} 
& 0.7025 \best{{\small(+0.01)}} \\

\hline

DP$^2$O-SR~\cite{wu2025dp2osr}
& 4.2159 
& 67.9658 
& 0.5703 
& 3.865 
& 0.6867 \\

+COMPASS     
& 4.5796 \best{{\small(+0.36)}} 
& 72.8912 \best{{\small(+4.93)}} 
& 0.6123 \best{{\small(+0.04)}} 
& 3.949 \best{{\small(+0.08)}} 
& 0.6982 \best{{\small(+0.01)}} \\

\shline
\end{tabular}}
\label{tab:nr_metrics_posttrain}
\vspace{-5pt}
\end{table}

\subsection{Comparison with DPO-based SR Methods}
We further compare our online SR method with the offline RL approach DP$^2$O-SR\cite{wu2025dp}, and also verify the generality of OARS on different backbones (e.g., Flux-ControlNet~\cite{blackforestlabs_flux2024}). All experiments are conducted on RealSet80. Noted that DP$^2$O-SR can be viewed as most closely related to our stage-1 full-reference alignment; however, since its training data and protocol are not publicly available, we cannot reproduce an exactly matched stage-1 setting, and a fully controlled end-to-end comparison is therefore non-trivial. To ensure a consistent and well-defined training recipe, we adopt our stage-2 RL pipeline with the COMPASS reward.

As shown in Tab.~\ref{tab:nr_metrics_posttrain}, applying our two-stage training yields great improvements on RealSet80 across multiple NR metrics (LIQE, MUSIQ, TOPIQ), and additional test-time tuning (TTT) further enhances the results, where ``TTT'' denotes an extra RL stage performed on the RealSet80 test set after completing the non-reference RL stage. Moreover, when we directly apply our non-reference RL stage on top of the DP$^2$O-SR model, we observe consistent gains across all metrics. This suggests that our online, process-aware optimization provides an additional refinement signal beyond offline DPO-style alignment. More backbone results are provided in the \textbf{supplementary materials}.

\begin{table}[t!]
\centering
\caption{\textbf{Ablation study on different RL stages and initialization settings}.}
\vspace{-10pt}
\renewcommand{\arraystretch}{1.15}
\resizebox{\linewidth}{!}{
\begin{tabular}{l c c c c c c c c c c c}
\shline
Method & RL Stage & SFT~/~Base 
& PSNR$\uparrow$ & SSIM$\uparrow$ 
& LPIPS$\downarrow$ & DISTS$\downarrow$ 
& LIQE$\uparrow$ & MUSIQ$\uparrow$ 
& MANIQA$\uparrow$ & Q-Insight$\uparrow$ & TOPIQ$\uparrow$ \\
\hline\hline

Qwen-SFT 
& - & - 
& \best{22.71} & \third{0.6462} & \secondbest{0.3100} & \best{0.2203} 
& 3.8146 & 68.5687 & 0.4897 & 3.545 & 0.6397 \\

case (1) 
& stage1 & base 
& \secondbest{22.52} & \best{0.6494} & \third{0.3115} & \secondbest{0.2224} 
& \secondbest{4.2345} & \secondbest{71.0159} & \third{0.5244} & 3.685 & 0.6741 \\

case (2) 
& stage1+stage2 & base 
& \third{22.36} & \secondbest{0.6481} & \best{0.3095} & \third{0.2244} 
& \best{4.3045} & \best{71.4136} & \secondbest{0.5277} & \best{3.701} & \best{0.6803} \\

case (3) 
& stage1 & sft 
& 22.15 & 0.6266 & 0.3249 & 0.2272 
& 4.0784 & \third{70.6030} & 0.5233 & \third{3.617} & \third{0.6756} \\

case (4) 
& stage1+stage2 & sft 
& 21.31 & 0.5956 & 0.3326 & 0.2334 
& \third{4.0937} & 70.5618 & \best{0.5293} & \secondbest{3.667} & \secondbest{0.6768} \\

\shline
\end{tabular}}
\label{tab:ablation_rl_stage}
\vspace{-15pt}
\end{table}

\section{Conclusion}
We presented OARS, a process-aware online alignment framework for generative real-world super-resolution. We first propose COMPASS, an MLLM-based reward that scores the LR$\rightarrow$SR transition by jointly modeling fidelity and perceptual gain with a quality-adaptive trade-off. To train it, we curate COMPASS-20K with diverse LR--SR pairs from multiple algorithms and produce fine-grained perceptual labels via a three-stage pipeline. Notably, COMPASS achieves state-of-the-art performance on SR preference benchmarks without ground-truth references. Guided by this reward, OARS performs progressive online alignment from cold-start flow matching to reference-based and finally reference-free RL, enabling on-policy exploration while reducing reward hacking and hallucinations. Experiments demonstrate that OARS provides an effective post-training paradigm for aligning generative SR models with human preferences.
\clearpage  

%
%
\bibliographystyle{splncs04}
\bibliography{main}





%


\clearpage
\appendix
\section{Appendix}
\subsection{Construction Details of COMPASS-20K}
For the real-world low-quality subset, we manually collected 1,600 color RGB images from publicly accessible online sources, with attention to image quality, source diversity, content diversity, and the original usage conditions of the source content. This subset is used solely for research and experimental purposes.

The collected images cover a wide range of common real-world degradations, including noise, compression artifacts, defocus blur, motion blur, and lens blur. All images have resolutions ranging from 360p to 480p.

In terms of content, the dataset spans diverse everyday and natural scenes, including portraits, food and baking, urban streets, shopping malls, vehicles, sports scenes, flowers, pets, landscapes, and natural environments such as wildlife, plants, and underwater creatures.

In addition to the real-world images, we further include 800 images from DIV2K. For these images, we apply random Real-ESRGAN-style\cite{wang2021real} degradations to synthesize low-resolution inputs.

To construct the LR--SR pairs, we apply 12 SR methods to each input image for 4× super-resolution, including DiffBIR\cite{lin2024diffbir}, FeMaSR\cite{chen2022femasr}, InvSR\cite{yue2025arbitrary}, OSEDiff\cite{wu2024one}, PASD\cite{yang2024pixel}, Real-ESRGAN\cite{wang2021real}, ResShift\cite{yue2024resshift}, S3Diff\cite{zhang2024degradation}, SeeSR\cite{wu2024seesr}, SinSR\cite{wang2024sinsr}, StableSR\cite{wang2024exploiting}, and SwinIR\cite{SwinIR}.

\begin{figure}[b!]
    \centering
    \includegraphics[width=\linewidth]{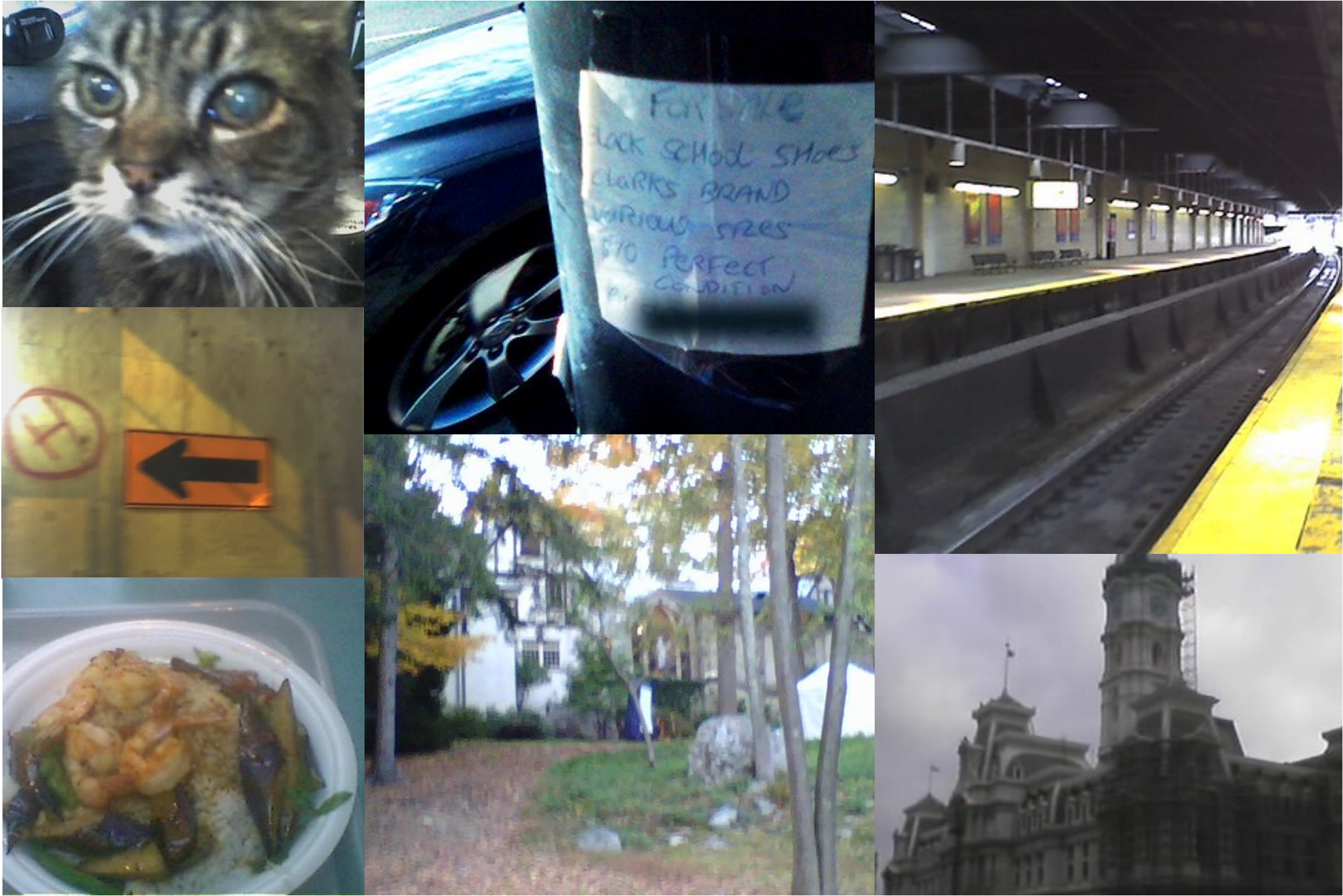}
    \caption{\textbf{Examples from the real-world low-quality subset of COMPASS-20K.}}
    \label{fig:real_lq_examples}
\end{figure}
\subsection{User Study Setup}
To further assess the effectiveness of OARS, we conduct a user study against several representative SR methods. We select 12 LR images from the RealSR, DRealSR, and DIV2K datasets for evaluation. The compared methods include DP$^2$O-SR\cite{wu2025dp2osr}, OSEDiff\cite{wu2024one}, and UARE\cite{li2025uare}. We invite 27 expert researchers to choose the best super-resolved/restored result for each sample based on two equally weighted criteria: (i) perceptual quality, focusing on clarity, details, and realism, and (ii) consistency with the LR input, including the alignment of image structure and texture. As stated in the main text, OARS receives 47.62\% of the votes, the highest proportion among all compared methods, surpassing the second-best method, DP$^2$O-SR, which receives 27.68\%.

\subsection{More Ablation Studies}
\paragraph{\textbf{Ablation on Reward Model on SR Task.}} To validate the effectiveness of the proposed COMPASS reward, we conduct controlled experiments on Qwen-SFT using the same RL framework while replacing COMPASS with alternative reward models, including RALI~\cite{zhao2025reasoning}, HPSv2, and Qwen25-VL-7B. For Qwen25-VL-7B, we follow a strategy similar to Edit-R1~\cite{lin2025uniworld} by using a logit-weighted aggregation as the reward signal.

As reported on Tab.~\ref{tab:overall_metrics}, we observe that general-purpose models such as HPSv2 and Qwen25-VL are unable to provide effective feedback for SR tasks. In particular, metrics including TOPIQ and MUSIQ even exhibit noticeable degradation. Compared to RALI~\cite{zhao2025reasoning}, which relies solely on perceptual quality signals without explicit fidelity constraints, these methods can yield certain perceptual gains but suffer from severe fidelity degradation relative to Qwen-SFT. In contrast, COMPASS effectively prevents the policy from exploiting reward signals that purely favor perceptual sharpness, which often leads to over-sharpening and over-saturation artifacts in super-resolved images. By explicitly balancing fidelity preservation and perceptual enhancement, COMPASS achieves a more stable and effective trade-off, making it particularly well suited for super-resolution.

\paragraph{\textbf{Comparison with Flow-GRPO.}}
Both Flow-GRPO~\cite{liu2025flow} and DiffusionNFT~\cite{zheng2025diffusionnft} aim to align flow-based generative models with reward models using reinforcement learning signals, but they differ in how the optimization signal is incorporated. Flow-GRPO performs trajectory-level RL on the reverse denoising process, treating each denoising step as a policy action and applying policy-gradient updates along the full sampling trajectory. While theoretically well aligned with RL optimization, this approach requires storing complete trajectories and computing likelihood ratios or importance weights, leading to high computational and memory costs. In contrast, DiffusionNFT performs RL on the forward diffusion training objective: it generates samples, scores them with a reward model, and converts the reward signal into a weighted diffusion loss through positive–negative sample pairs. This formulation avoids trajectory storage and likelihood-ratio computation, resulting in significantly higher training efficiency.

For Real-ISR task, DiffusionNFT provides two key advantages over Flow-GRPO. First, its training procedure is significantly more efficient: in practice we observe 5–10$\times$ faster training and convergence compared with trajectory-level RL approaches such as Flow-GRPO. Second, super-resolution is a strongly conditioned image editing task, where the high-resolution output must remain tightly consistent with the low-resolution input. Unlike open-ended image generation, the solution space is already heavily constrained, and extensive exploration of the generative trajectory is unnecessary. In this context, introducing additional stochasticity by converting deterministic flow ODEs into stochastic SDEs, as typically required in trajectory-level RL optimization, may unnecessarily increase randomness in the generation process. Such stochasticity can exacerbate the well-known perceptual–fidelity trade-off in super-resolution, leading to instability between reconstruction accuracy and perceptual quality. Based on DiffusionNFT, our OARS provides a more stable and efficient mechanism for incorporating reward feedback under strong conditional constraints.

Further, we compare Flow-GRPO and our NFT-based OARS framework on the super-resolution task. We adopt the default configuration of Flow-GRPO and use two typical rewards, RALI and our COMPASS. As shown in Table~\ref{tab:overall_metrics}, we find that the performance of Flow-GRPO is lower than RL methods using NFT across multiple metrics, especially on NR metrics that reflect image quality. This indicates that Flow-GRPO suffers from low convergence efficiency and optimization difficulty. Our OARS performs multi-stage, forward-process-based RL, leading to more stable training and better alignment with human perception.

\paragraph{\textbf{Ablation on More Generative backbones.}}We further apply the OARS framework to additional generative models, such as Flux~\cite{blackforestlabs_flux2024}, to evaluate its generality. Specifically, we first train Flux on the same dataset using flow matching loss (dubbed Flux-SFT) and then perform the same two-stage RL training. As reported in Table~\ref{tab:oars_results}, the model trained with OARS achieves significant improvements in perceptual quality while maintaining comparable fidelity. In particular, no-reference metrics such as MANIQA and TOPIQ show notable gains. These results demonstrate the strong generalization ability of our RL framework.

\begin{figure*}[b!]
    \centering
    \includegraphics[width=\textwidth]{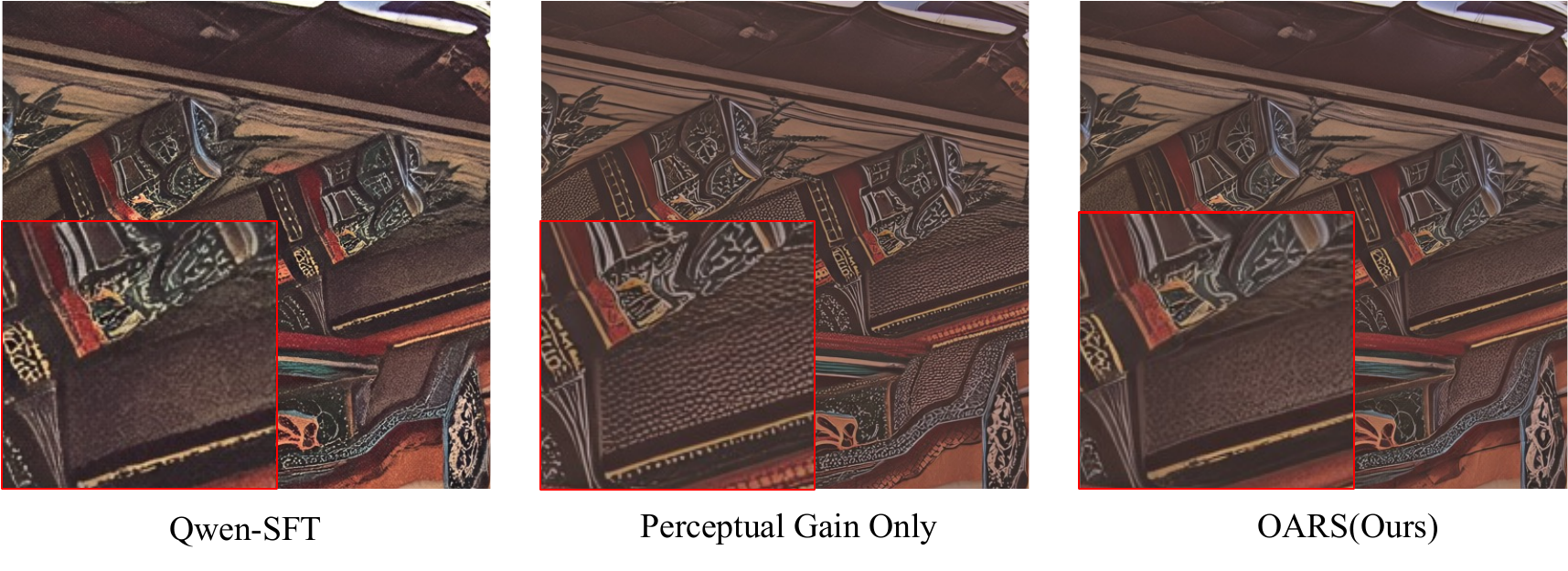}
    \caption{\textbf{Effect of perceptual-gain-only optimization.} Optimizing the policy using only $\Delta Q = Q_{\mathrm{SR}} - Q_{\mathrm{LR}}$ as the RL reward leads to severe artifacts and poor consistency with the input image, revealing clear reward hacking behavior. In contrast, COMPASS preserves fidelity while achieving clear perceptual quality improvement.}    \label{fig:rewardhacking}
\end{figure*}

\paragraph{\textbf{Ablation on More Reward Formulations}}
In the main paper, we have already shown on the SR preference benchmark that our reward design aligns well with human visual preference, and we further ablate its core formulation. To more directly verify the effectiveness of the reward formulation during reinforcement learning, we present additional comparisons with alternative reward designs in Table~\ref{tab:ablation_reward_formulation}. When only the perceptual gain term is used as the reward, the full-reference metrics drop substantially. Although the no-reference metrics improve, this improvement is largely a result of reward hacking rather than genuine visual enhancement. As shown in Fig.~\ref{fig:rewardhacking}, without an explicit fidelity constraint, the model tends to produce unreal artifacts that artificially boost the no-reference scores. Introducing a gating term, \ie, $F^{Q_{\mathrm{LR}}/\gamma}\cdot \Delta Q$ (Formulation 2), alleviates reward hacking on no-reference metrics and mitigates the severe fidelity drop. By further incorporating an input-quality-aware fidelity term, our final COMPASS formulation is able to achieve perceptual quality gains while preserving reconstruction fidelity.

\begin{table}[t]
\centering
\caption{\textbf{Ablation Study on different generative backbones.} $\uparrow$ / $\downarrow$ indicate higher / lower is better.}
\resizebox{1.0\linewidth}{!}{
\begin{tabular}{lcccccccc}
\shline
Method & PSNR$\uparrow$ & SSIM$\uparrow$ & LPIPS$\downarrow$ & DISTS$\downarrow$ & LIQE$\uparrow$ & MUSIQ$\uparrow$ & MANIQA$\uparrow$ & TOPIQ$\uparrow$ \\
\hline \hline
Qwen-SFT       & \third{22.71} & 0.6462 & \secondbest{0.3100} & \third{0.2203} & 3.8146 & \secondbest{68.5687} & \third{0.4897} & \third{0.6397} \\
Qwen-SFT+OARS  & 22.36 & \third{0.6481} & \best{0.3095} & 0.2244 & \best{4.3045} & \best{71.4136} & \best{0.5277} & \best{0.6803} \\
Flux-SFT          & \best{23.62}   & \best{0.6617} & \third{0.3224} & \best{0.2168} & \third{3.8442} & 67.1802 & 0.4686 & 0.6031 \\
Flux+OARS      & \secondbest{23.10}   & \secondbest{0.6603} & 0.3335 & \secondbest{0.2174} & \secondbest{4.1277} & \third{68.3608} & \secondbest{0.5035} & \secondbest{0.6427} \\
\shline
\end{tabular}}
\label{tab:oars_results}
\end{table}

\begin{table}[t!]
\centering
\caption{\textbf{Quantitative comparison of different methods under full-reference and no-reference image quality assessment metrics.} $\uparrow$ / $\downarrow$ indicates higher / lower is better.}
\vspace{-10pt}
\resizebox{1.0\linewidth}{!}{
\renewcommand{\arraystretch}{1.15}
\begin{tabular}{l | c | c c c c c c c c}
\shline
Method 
& RL Method
& PSNR$\uparrow$ 
& SSIM$\uparrow$ 
& LPIPS$\downarrow$ 
& DISTS$\downarrow$ 
& LIQE$\uparrow$ 
& MUSIQ$\uparrow$ 
& MANIQA$\uparrow$ 
& TOPIQ$\uparrow$ \\
\hline \hline

Qwen-SFT 
& --
& \third{22.71} 
& 0.6462 
& \third{0.3100} 
& \secondbest{0.2203} 
& 3.8146 
& 68.5687 
& 0.4897 
& 0.6397 \\

+HPSv2    
& OARS
& \secondbest{22.91} 
& \secondbest{0.6547} 
& 0.3154 
& \third{0.2238} 
& 3.8294 
& 68.6830 
& 0.4966 
& 0.6383 \\

+Qwen25   
& OARS
& \best{23.08} 
& \best{0.6618} 
& \best{0.3042} 
& \best{0.2197} 
& 3.7682 
& 68.2220 
& 0.4835 
& 0.6273 \\ 

+RALI     
& OARS
& 20.83 
& 0.5283 
& 0.3777 
& 0.2472 
& \secondbest{3.9657} 
& \secondbest{71.0329} 
& \best{0.5399} 
& \secondbest{0.6732} \\

+COMPASS (Ours)     
& OARS
& 22.36 
& \third{0.6481} 
& \secondbest{0.3095} 
& 0.2244 
& \best{4.3045} 
& \best{71.4136} 
& \secondbest{0.5277} 
& \best{0.6803} \\

\hline \hline

+RALI   
& Flow-GRPO
& 21.08	
& 0.5781	
& 0.3548
& 0.2379	
& 3.8757	
& \third{70.3903}	
& \third{0.5127}
& \third{0.6639}  \\

+COMPASS   
& Flow-GRPO
& 21.82
& 0.6150
& 0.3446
& 0.2325 
& \third{3.9351}
& 69.8253
& 0.5011
& 0.6527 \\

\shline
\end{tabular}}
\label{tab:overall_metrics}
\end{table}

\begin{table}[t!]
\centering
\caption{\textbf{Ablation on reward formulation.} Formulation 1 uses only perceptual gain, $\Delta Q = Q_{\mathrm{SR}} - Q_{\mathrm{LR}}$, while Formulation 2 adds a gating term, $F^{Q_{\mathrm{LR}}/\gamma}\!\cdot\!\Delta Q$. COMPASS further introduces an input-quality-aware fidelity term, enabling perceptual improvement with better fidelity preservation.}
\vspace{-10pt}
\resizebox{1.0\linewidth}{!}{
\renewcommand{\arraystretch}{1.15}
\begin{tabular}{l | c c c c c c c c}
\shline
Method 
& PSNR$\uparrow$ 
& SSIM$\uparrow$ 
& LPIPS$\downarrow$ 
& DISTS$\downarrow$ 
& LIQE$\uparrow$ 
& MUSIQ$\uparrow$ 
& MANIQA$\uparrow$ 
& TOPIQ$\uparrow$ \\
\hline \hline

Qwen-SFT
& 22.71
& 0.6462
& 0.3100
& 0.2203
& 3.8146
& 68.5687
& 0.4897
& 0.6397 \\

+Formulation 1
& 21.38
& 0.6039
& 0.3436
& 0.2409
& 4.3654
& 71.8348
& 0.5596
& 0.7058 \\

+Formulation 2
& 21.89
& 0.6203
& 0.3321
& 0.2320
& 4.2445
& 70.9428
& 0.5426
& 0.6904 \\

+COMPASS (Ours)
& 22.36
& 0.6481
& 0.3095
& 0.2244
& 4.3045
& 71.4136
& 0.5277
& 0.6803 \\

\shline
\end{tabular}}
\label{tab:ablation_reward_formulation}
\vspace{-20pt}
\end{table}

\begin{table*}[b]
\centering
\caption{\textbf{Quantitative comparison on four SR datasets}, including RealSR, DRealSR, DIV2K, and RealSet80. 
$\uparrow/\downarrow$ indicates higher/lower is better.}
\label{tab:more_quantitative}
\vspace{-10pt}
\renewcommand{\arraystretch}{1.15}
\resizebox{\linewidth}{!}{
\begin{tabular}{c | l | c c c c | c c c c c}
\hline
Dataset & Method
& PSNR$\uparrow$ & SSIM$\uparrow$ & LPIPS$\downarrow$ & DISTS$\downarrow$
& LIQE$\uparrow$ & MUSIQ$\uparrow$ & MANIQA$\uparrow$ & Q-Insight$\uparrow$ & TOPIQ$\uparrow$ \\
\hline\hline

\multirow{12}{*}{\textbf{RealSR}}
& DiffBIR~\cite{lin2024diffbir} & 23.20 & 0.6346 & 0.3350 & 0.2162 & 3.5529 & 65.25 & 0.4620 & 3.530 & 0.6033 \\
& OSEDiff~\cite{wu2024one} & 23.07 & 0.6850 & 0.2941 & 0.2109 & \secondbest{4.0681} & \third{68.95} & 0.4876 & \best{3.712} & \third{0.6441} \\
& PASD~\cite{yang2024pixel} & \best{24.50} & \secondbest{0.7115} & \best{0.2716} & \best{0.1954} & 2.8541 & 58.52 & 0.3831 & 3.173 & 0.4969 \\
& ResShift~\cite{yue2024resshift} & \third{24.17} & 0.6528 & 0.4336 & 0.2812 & 2.6610 & 53.38 & 0.3412 & 2.781 & 0.4210 \\
& S3Diff~\cite{zhang2024degradation} & 23.16 & 0.6810 & \secondbest{0.2748} & \secondbest{0.1986} & 4.0080 & 67.57 & 0.4677 & \third{3.674} & 0.6301 \\
& SeeSR~\cite{wu2024seesr} & \secondbest{24.34} & \best{0.7187} & \third{0.2754} & 0.2134 & 3.3938 & 65.53 & 0.4856 & 3.285 & 0.6246 \\
& SinSR~\cite{wang2024sinsr} & 23.68 & 0.6649 & 0.3490 & 0.2445 & 3.2255 & 61.03 & 0.4230 & 3.264 & 0.5383 \\
& StableSR~\cite{wang2024exploiting} & 23.73 & \third{0.6979} & 0.2792 & \third{0.2023} & 3.0532 & 61.65 & 0.3826 & 3.480 & 0.5201 \\
& PURE~\cite{wei2025perceive} & 21.31 & 0.5738 & 0.3859 & 0.2468 & 3.7881 & 66.57 & 0.4829 & 3.569 & 0.6301 \\
& UARE~\cite{li2025uare} & 21.38 & 0.6464 & 0.3095 & 0.2344 & \third{4.0658} & \secondbest{69.67} & \secondbest{0.5260} & 3.664 & \secondbest{0.6796} \\
& Qwen-SFT~\cite{wu2025qwen} & 22.71 & 0.6462 & 0.3100 & 0.2203 & 3.8146 & 68.57 & \third{0.4897} & 3.545 & 0.6397 \\
& OARS (Ours) & 22.36 & 0.6481 & 0.3095 & 0.2244 & \best{4.3045} & \best{71.41} & \best{0.5277} & \secondbest{3.701} & \best{0.6803} \\

\hline\hline

\multirow{12}{*}{\textbf{DRealSR}}
& DiffBIR~\cite{lin2024diffbir} & 26.08 & 0.6578 & 0.4144 & 0.2564 & 3.3993 & 61.81 & 0.4612 & 3.397 & 0.6084 \\
& OSEDiff~\cite{wu2024one} & 25.60 & 0.7403 & 0.3088 & 0.2158 & \third{3.9797} & \third{65.24} & \third{0.4879} & \secondbest{3.600} & \third{0.6273} \\
& PASD~\cite{yang2024pixel} & \secondbest{28.18} & 0.7722 & 0.2970 & 0.2108 & 2.6129 & 51.42 & 0.3595 & 2.746 & 0.4587 \\
& ResShift~\cite{yue2024resshift} & 27.39 & 0.6907 & 0.4996 & 0.3077 & 1.7905 & 40.58 & 0.2457 & 2.354 & 0.3414 \\
& S3Diff~\cite{zhang2024degradation} & 26.18 & 0.7197 & 0.3161 & \secondbest{0.2099} & 3.9255 & 63.34 & 0.4635 & 3.488 & 0.6181 \\
& SeeSR~\cite{wu2024seesr} & \third{28.14} & \secondbest{0.7798} & \secondbest{0.2832} & 0.2241 & 2.7943 & 55.89 & 0.3976 & 2.891 & 0.5436 \\
& SinSR~\cite{wang2024sinsr} & 26.72 & 0.6933 & 0.4031 & 0.2624 & 2.7781 & 53.36 & 0.3677 & 3.000 & 0.4959 \\
& StableSR~\cite{wang2024exploiting} & \best{28.28} & \best{0.7981} & \best{0.2687} & \best{0.2026} & 2.5068 & 51.62 & 0.3226 & 2.736 & 0.4355 \\
& PURE~\cite{wei2025perceive} & 23.04 & 0.5718 & 0.4461 & 0.2674 & 3.7390 & 60.68 & 0.4362 & 3.286 & 0.5888 \\
& UARE~\cite{li2025uare} & 21.31 & 0.5736 & 0.4071 & 0.2613 & \secondbest{4.0445} & \secondbest{67.71} & \best{0.5121} & \best{3.651} & \best{0.6652} \\
& Qwen-SFT~\cite{wu2025qwen} & 24.60 & 0.6623 & 0.3430 & 0.2288 & 3.5585 & 62.96 & 0.4348 & 3.313 & 0.5835 \\
& OARS (Ours) & 24.35 & 0.6750 & 0.3319 & 0.2347 & \best{4.2088} & \best{67.97} & \secondbest{0.4927} & \third{3.519} & \secondbest{0.6438} \\

\hline\hline

\multirow{12}{*}{\textbf{DIV2K}}
& DiffBIR~\cite{lin2024diffbir}  & 18.94 & 0.4332 & 0.4009 & 0.2238 & 3.8573 & 67.20 & 0.4574 & 3.577 & 0.6467 \\
& OSEDiff~\cite{wu2024one}  & 18.86 & \third{0.4563} & \secondbest{0.3579} & 0.2209 & 3.8877 & 67.83 & 0.4422 & 3.722 & 0.6269 \\
& PASD~\cite{yang2024pixel} & 18.98 & 0.4562 & 0.4293 & 0.2373 & 3.6022 & 63.46 & 0.4025 & 3.410 & 0.5653 \\
& ResShift~\cite{yue2024resshift} & \secondbest{19.15} & 0.4311 & 0.4900 & 0.2808 & 2.8862 & 56.02 & 0.3534 & 2.924 & 0.4662 \\
& S3Diff~\cite{zhang2024degradation} & 18.76 & 0.4490 & \best{0.3299} & \best{0.1990} & 4.2692 & 69.31 & 0.4675 & 3.760 & 0.6679 \\
& SeeSR~\cite{wu2024seesr} & \third{19.11} & \secondbest{0.4580} & \third{0.3769} & 0.2339 & 3.7445 & 66.31 & 0.4686 & 3.611 & 0.6330 \\
& SinSR~\cite{wang2024sinsr} & 18.58 & 0.4059 & 0.4483 & 0.2455 & 3.4629 & 64.12 & 0.4483 & 3.224 & 0.5997 \\
& StableSR~\cite{wang2024exploiting} & \best{19.85} & \best{0.4940} & 0.4796 & 0.2887 & 1.8466 & 43.25 & 0.2181 & 2.066 & 0.3276 \\
& PURE~\cite{wei2025perceive} & 16.71 & 0.3661 & 0.4449 & 0.2293 & \third{4.2701} & 70.06 & \best{0.5201} & 3.721 & 0.6621 \\
& UARE~\cite{li2025uare} & 16.59 & 0.3857 & 0.4074 & 0.2138 & 4.2627 & \third{70.45} & \third{0.5028} & \secondbest{3.823} & \secondbest{0.6864} \\
& Qwen-SFT~\cite{wu2025qwen} & 17.31 & 0.3988 & 0.3811 & \secondbest{0.1993} & \secondbest{4.3404} & \secondbest{72.35} & 0.4875 & \third{3.818} & \third{0.6745} \\
& OARS (Ours) & 17.32 & 0.4072 & 0.3812 & \third{0.2036} & \best{4.6668} & \best{74.07} & \secondbest{0.5052} & \best{3.940} & \best{0.6960} \\

\hline\hline

\multirow{9}{*}{\textbf{RealSet80}}
& DiffBIR~\cite{lin2024diffbir} & \multicolumn{4}{c|}{} & 4.1113 & 68.10 & \best{0.5527} & 3.684 & \third{0.6736} \\
& OSEDiff~\cite{wu2024one} & \multicolumn{4}{c|}{} & 4.2251 & 68.88 & 0.4995 & 3.725 & 0.6062 \\
& SeeSR~\cite{wu2024seesr} & \multicolumn{4}{c|}{} & \secondbest{4.3317} & 69.70 & 0.5362 & \secondbest{3.774} & \secondbest{0.6887} \\
& SinSR~\cite{wang2024sinsr} & \multicolumn{4}{c|}{} & 3.6613 & 62.78 & 0.4483 & 3.382 & 0.5854 \\
& StableSR~\cite{wang2024exploiting} & \multicolumn{4}{c|}{\Large N/A} & 3.9074 & 67.67 & 0.4682 & 3.562 & 0.6440 \\
& PURE~\cite{wei2025perceive} & \multicolumn{4}{c|}{} & \third{4.2528} & 69.55 & 0.5215 & \third{3.744} & 0.6647 \\
& UARE~\cite{li2025uare} & \multicolumn{4}{c|}{} & 4.1804 & \secondbest{70.05} & \third{0.5363} & 3.508 & 0.6446 \\
& Qwen-SFT~\cite{wu2025qwen} & \multicolumn{4}{c|}{} & 4.1602 & \third{69.79} & 0.5171 & 3.701 & 0.6539 \\
& OARS (Ours) & \multicolumn{4}{c|}{} & \best{4.5465} & \best{72.96} & \secondbest{0.5364} & \best{3.823} & \best{0.6967} \\

\hline
\end{tabular}}
\label{tab:maintable_complete}
\vspace{-25pt}
\end{table*}

\subsection{More Quantitative Comparison}
We further expand the quantitative comparison in Tab.~\ref{tab:more_quantitative} by including more super-resolution baselines and additional results on DRealSR. On the one hand, OARS consistently achieves state-of-the-art performance on subjective quality metrics compared with previous methods. On the other hand, even when compared with methods that already obtain strong perceptual results, such as UARE and PURE, OARS still shows clear advantages on full-reference metrics. These results indicate that OARS is able to improve perceptual quality while maintaining strong fidelity. In addition, compared with Qwen-SFT, OARS significantly improves subjective metrics without compromising consistency, which further validates the effectiveness of the proposed alignment framework.
\subsection{More Qualitative Comparison}
\begin{figure*}[p]
    \centering
    \includegraphics[width=\textwidth]{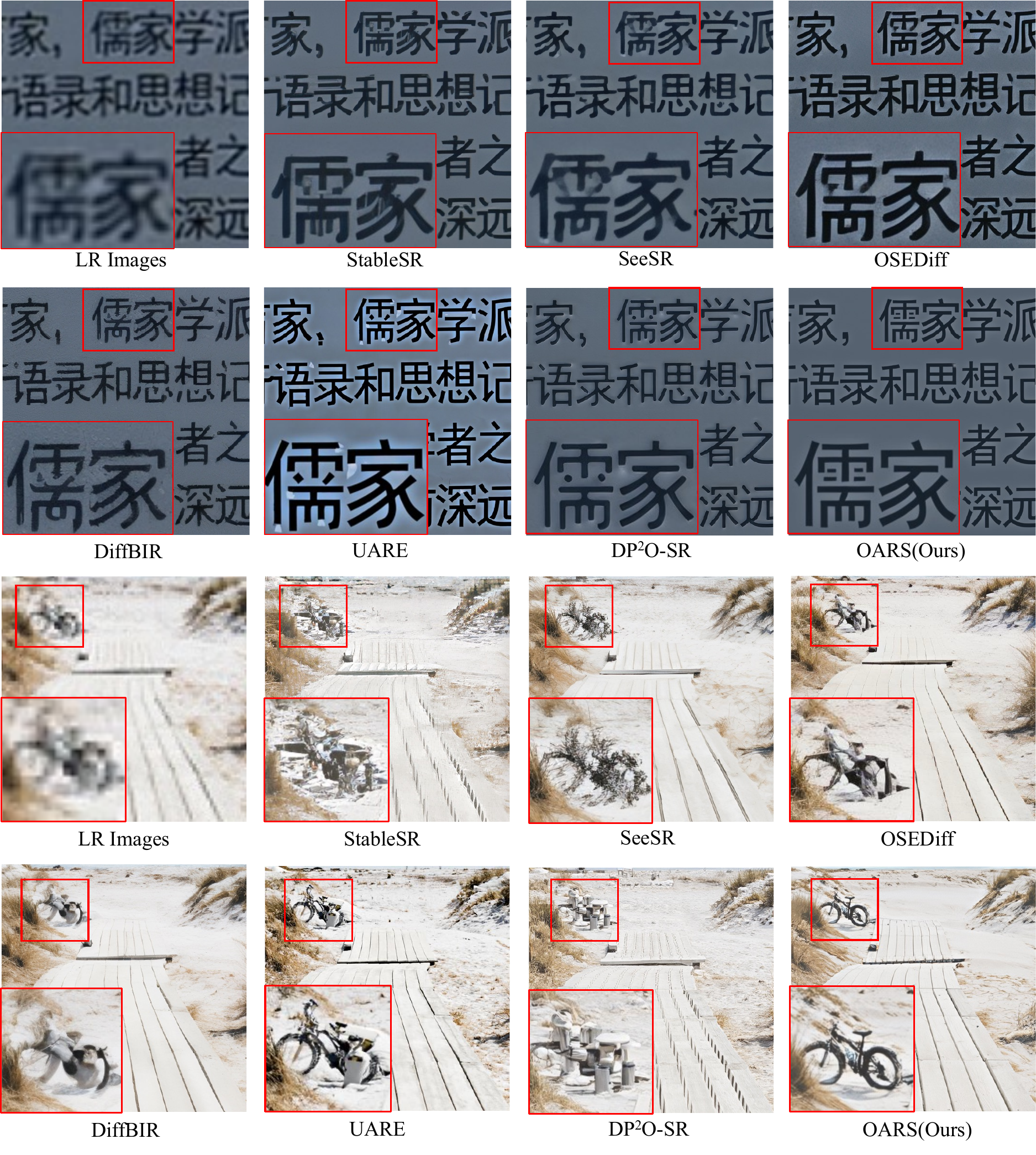}
    \caption{More qualitative comparisons of OARS against state-of-the-art Real-ISR methods under complex degradations.}
    \label{fig:appendix_qual1}
\end{figure*}

\begin{figure*}[p]
    \centering
    \includegraphics[width=\textwidth]{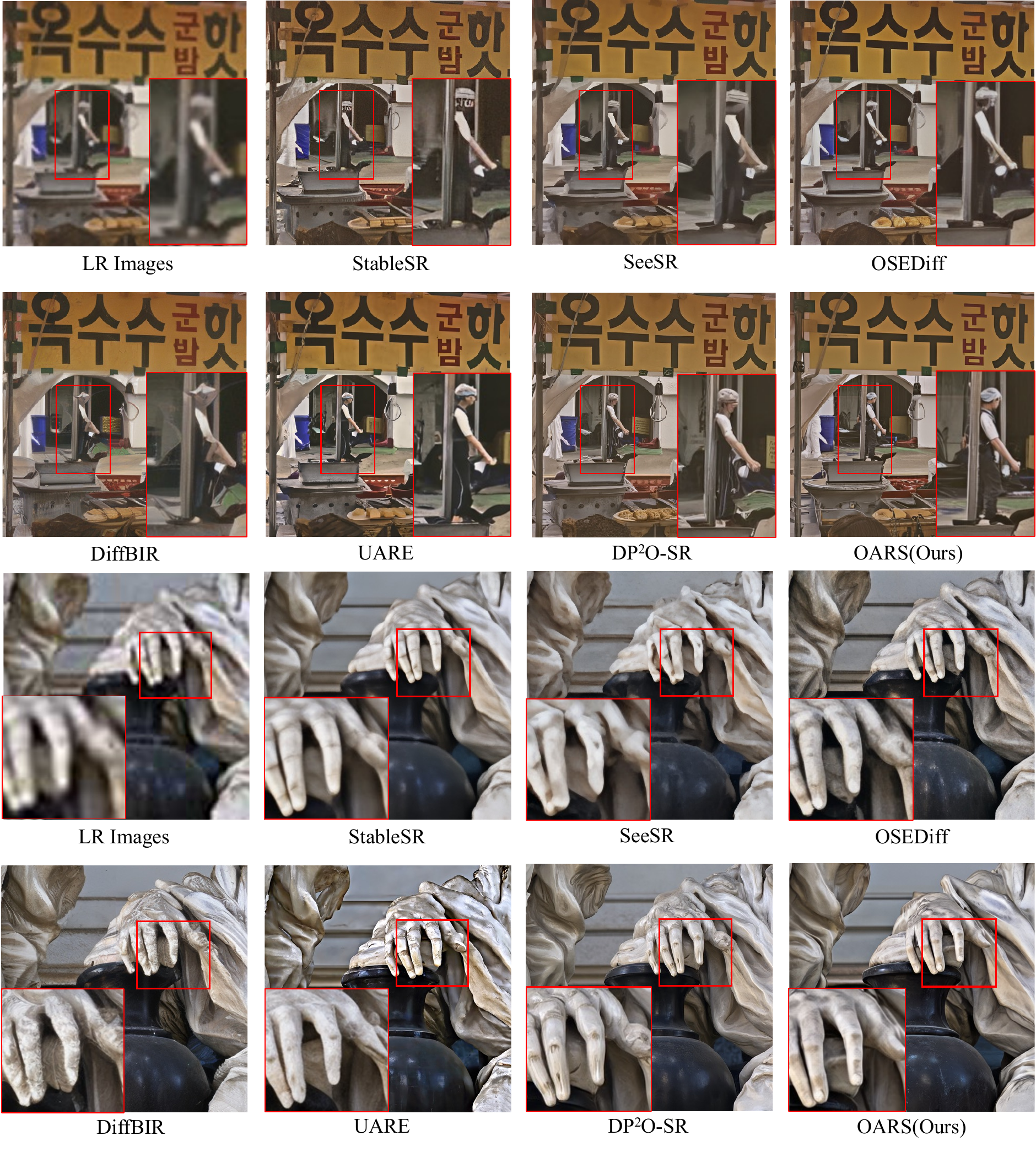}
    \caption{More qualitative comparisons of OARS against state-of-the-art Real-ISR methods under complex degradations.}
    \label{fig:appendix_qual2}
\end{figure*}

To further highlight the subjective merits of our method, we present additional qualitative comparisons of OARS against state-of-the-art Real-ISR approaches in Fig.~\ref{fig:appendix_qual1} and Fig.~\ref{fig:appendix_qual2}. The examples are drawn from RealSR, DRealSR, DIV2K, and RealSet. The results show that OARS achieves superior perceptual quality while better preserving reconstruction fidelity than the compared methods.
\subsection{Limitations}
First, our reward relies on an MLLM-based evaluator, which incurs substantial computational overhead during online reinforcement learning and limits training efficiency and scalability. A natural direction is to adopt lighter-weight reward models, as in RALI\cite{zhao2025reasoning}, or distill the MLLM reward into a smaller scorer. Second, the current framework is designed for image only. Extending the proposed reward design and online RL framework to video task, where temporal consistency becomes critical, is an important direction for future work.

\clearpage


\end{document}